\newif\ifarxiv
\arxivtrue 

\ifarxiv
    \documentclass[%
     amsmath,amssymb,%nobibnotes,
     preprint,%
    author-year,%
    ]{revtex4-2}
  \usepackage{array}[=2016-10-06]
  \usepackage{amsmath,amssymb,graphicx}
  \makeatletter
    \providecommand\@captype{figure}
  \makeatother

  \newcommand{\orcid}[1]{ \href{https://orcid.org/#1}{(ORCID: #1)}}
  \newcommand{\affil}[1]{\affiliation{#1}}
  \newcommand{\ack}[1]{\begin{acknowledgments}#1\end{acknowledgments}}
\else
  \documentclass{iopjournal}
\fi

\usepackage[caption=false]{subfig}

\usepackage[T1]{fontenc}
\usepackage[utf8]{inputenc}
\usepackage[english]{babel}

\usepackage{graphicx}
\usepackage{enumitem}
\usepackage{textcomp}
\usepackage{gensymb}
\usepackage{xparse}
\usepackage{siunitx}
\usepackage{isotope}
\usepackage{rotating}
\usepackage{multirow}

\ifarxiv
\else
  \usepackage[sort]{natbib}
\fi
\ifarxiv
  \setcitestyle{authoryear, round}
\else
  \setcitestyle{authoryear, round}
\fi

\expandafter\let\csname equation*\endcsname=\relax
\expandafter\let\csname endequation*\endcsname=\relax
\usepackage{amsmath}

\usepackage{amsfonts,amssymb,amscd}
\usepackage{xcolor, soul}
\usepackage{xspace}
\usepackage{microtype}

\usepackage[section]{placeins}

\sethlcolor{yellow}

\graphicspath{{figures/}}

\usepackage{acro}
\acsetup{single}
\DeclareAcronym{PET}{short=PET,long=Positron Emission Tomography}
\DeclareAcronym{LAFOV}{short=LAFOV,long=Long Axial Field of View}
\DeclareAcronym{FOV}{short=FOV,long=Field of View}
\DeclareAcronym{LOR}{short=LOR,long=Line of Response}
\DeclareAcronym{TOF}{short=TOF,long=time-of-flight}
\DeclareAcronym{CRT}{short=CRT,long=Coincidence Resolving Time}
\DeclareAcronym{CT}{short=CT,long=Computed Tomography}
\DeclareAcronym{AF}{short=AF,long=Attenuation Factor}
\DeclareAcronym{MC}{short=MC,long=Monte Carlo}
\DeclareAcronym{GATE}{short=GATE,long=Geant4 Application for Tomographic Emission}
\DeclareAcronym{NEMA}{short=NEMA IEC,long=NEMA IEC PET}
\DeclareAcronym{XCAT}{short=XCAT,long=4D eXtended CArdiac-Torso}
\DeclareAcronym{ML}{short=ML,long=Machine Learning}
\DeclareAcronym{BDT}{short=BDT,long=Boosted Decision Tree}
\DeclareAcronym{ANN}{short=ANN,long=Artificial Neural Network}
\DeclareAcronym{NN}{short=NN,long=Neural Network}
\DeclareAcronym{MLP}{short=MLP,long=Multi-Layer Perceptron}
\DeclareAcronym{XGBoost}{short=XGBoost,long=eXtreme Gradient Boosting}
\DeclareAcronym{AdaBoost}{short=AdaBoost,long=Adaptive Boosting}
\DeclareAcronym{GT}{short=GT,long=Ground Truth}
\DeclareAcronym{FWHM}{short=FWHM,long=full width at half maximum}
\DeclareAcronym{ROC}{short=ROC,long=receiver operating characteristic}
\DeclareAcronym{MCC}{short=MCC,long=Matthews correlation coefficient}
\DeclareAcronym{MSE}{short=MSE,long=Mean Squared Error}
\DeclareAcronym{CASToR}{short=CASToR,long=Customizable and Advanced Software for Tomographic Reconstruction}
\DeclareAcronym{MLEM}{short=MLEM,long=Maximum-Likelihood Expectation-Maximization}
\DeclareAcronym{FDG}{short=FDG,long=fluorodeoxyglucose}
\DeclareAcronym{CART}{short=CART,long=Classification and Regression tree}

\usepackage{tikz}
\usetikzlibrary{shapes.geometric, shapes.misc, arrows.meta, calc, positioning}
\usepackage{xcolor}
 
\tikzset{
  scanner/.style={
    draw, fill=white, rectangle,
    minimum width=2.5pt, minimum height=8pt,
    inner sep=0pt, line width=0.5pt
  },
  body/.style={
    draw=none, fill=blue!45!gray!80, circle
  },
  emissionpt/.style={
    draw=none, fill=red, circle, minimum size=5pt, inner sep=0pt
  },
  scatterpt/.style={
    draw=black, fill=white, circle, minimum size=8pt, inner sep=0pt, line width=0.9pt
  },
  startstop/.style={
    rounded rectangle, draw, fill=blue!20,
    minimum width=3.2cm, minimum height=0.9cm,
    align=center, font=\small
  },
  decision/.style={
    diamond, draw, fill=white,
    minimum width=2.8cm, minimum height=1.8cm,
    inner sep=1pt, align=center, font=\small,
    aspect=2.2
  },
  result/.style={
    rounded rectangle, draw,
    minimum width=2.4cm, minimum height=0.8cm,
    align=center, font=\small\bfseries
  },
  arrow/.style={-{Stealth[length=6pt]}, thick}
}
 
\newcommand{\drawstar}[3]{%
  \fill[yellow!85!orange, draw=black, line width=0.3pt]
    ({#1 + #3*sin(0)},   {#2 + #3*cos(0)})   --
    ({#1 + 0.4*#3*sin(36)},  {#2 + 0.4*#3*cos(36)})  --
    ({#1 + #3*sin(72)},  {#2 + #3*cos(72)})  --
    ({#1 + 0.4*#3*sin(108)}, {#2 + 0.4*#3*cos(108)}) --
    ({#1 + #3*sin(144)}, {#2 + #3*cos(144)}) --
    ({#1 + 0.4*#3*sin(180)},{#2 + 0.4*#3*cos(180)}) --
    ({#1 + #3*sin(216)}, {#2 + #3*cos(216)}) --
    ({#1 + 0.4*#3*sin(252)},{#2 + 0.4*#3*cos(252)}) --
    ({#1 + #3*sin(288)}, {#2 + #3*cos(288)}) --
    ({#1 + 0.4*#3*sin(324)},{#2 + 0.4*#3*cos(324)}) --
    cycle;
}
 
\newcommand{\scannerring}[4]{%
  \foreach \k in {1,...,#4}{%
    \pgfmathsetmacro{\ang}{(\k-1)*360/#4}%
    \node[scanner, rotate=\ang] at
      ({#1 + #3*cos(\ang)}, {#2 + #3*sin(\ang)}) {};%
  }%
  }

\usepackage{hyperref}
\usepackage{cleveref} % cleveref must be loaded after hyperref

\hypersetup{colorlinks, linkcolor={red!50!black}, citecolor={blue!50!black}, urlcolor={blue!80!black}}
\microtypesetup{
	protrusion=alltext-nott,
	expansion=alltext-nott,
	final
}

\begin{document}

\ifarxiv
%nothing
  \setcitestyle{round}
\else
    \articletype{Article type} %	 e.g. Paper, Letter, Topical Review...
\fi

\title{A systematic evaluation of machine learning classifiers for event-by-event background rejection in LAFOV PET scanners}

%\author{Author Name$^1$\orcid{0000-0000-0000-0000}, Author Name$^2$\orcid{0000-0000-0000-0000} and Author Name$^{1,*}$\orcid{0000-0000-0000-0000}}
%
%
%\affil{$^1$Department, Institution, City, Country}
%\affil{$^2$Department, Institution, City, Country}
%\affil{$^*$Author to whom any correspondence should be addressed.}

\author{
Konrad Klimaszewski$^{1,*}$\orcid{0000-0003-0741-5922},
Michał Obara$^1$\orcid{0009-0006-4779-6757},
Mateusz Bala$^1$\orcid{0000-0002-2505-7949},
Beatrix C. Hiesmayr$^{2,3}$\orcid{0000-0001-9062-6039},
Lech Raczyński$^1$\orcid{0000-0002-7039-2084},
Roman Y. Shopa$^1$\orcid{0000-0002-1089-5050},
Wojciech Zdeb$^1$\orcid{0009-0000-1485-251X},
Wojciech Krzemien$^{4,*}$\orcid{0000-0002-9546-358X}
}

\affil{$^1$National Centre for Nuclear Research, Department of Complex Systems, 05-400 Otwock, Poland}
\affil{$^2$IT:U Interdisciplinary Transformation University, Freistädter Strasse 400, 4040 Linz, Austria}
\affil{$^3$University of Vienna, Faculty of Physics, Währingerstrasse 17, 1090 Vienna, Austria}
\affil{$^4$National Centre for Nuclear Research, Department of High Energy Physics, 05-400 Otwock, Poland}

\email{konrad.klimaszewski@ncbj.gov.pl and wojciech.krzemien@ncbj.gov.pl}

\keywords{sample term, sample term, sample term}

\begin{abstract}
%%% DO NOT USE \ac{} IN THE ABSTRACT - HAS TO BE SELF CONTAINED - THE TEXT SHOULD ALSO BE INDEPENDENT
The introduction of Long Axial Field-of-View PET scanners brings significant sensitivity gains but also a substantial increase in the background rate from accidental coincidences, phantom-scattered and detector-scattered photons. While machine learning methods have been applied to background reduction in PET imaging, existing approaches target specific background components in post-processing rather than performing event-by-event classification on the raw data.
%%%
In this work, we formulate coincidence classification as a supervised multi-class problem and evaluate Boosted Decision Tree (XGBoost, AdaBoost) and Artificial Neural Network classifiers as pre-reconstruction filters, using Monte Carlo simulations of the Siemens Biograph Vision Quadra scanner with NEMA IEC and anthropomorphic XCAT phantoms. We investigate two feature sets: a 4-feature representation based on the Attenuation Factor (AF), photon time difference (dt), energy sum, and energy difference, and an extended 6-feature set that additionally incorporates topology-based variables.
%%%
A systematic robustness study via cross-phantom inference reveals that the 4-feature models generalise significantly better across different phantom geometries, with XGBoost suffering an accuracy loss of only 0.04 compared to 0.13 for the 6-feature variant. The Attenuation Factor and time difference are identified as the most discriminating features across all models and phantoms. Our best models achieve accuracies of up to 0.74 and 0.69 for the NEMA IEC and XCAT phantoms, respectively, outperforming traditional geometry-based cuts.
However, we show that this compact feature set not only provides limited suppression of in-phantom scattered coincidences, but it also can lead to non-trivial spatial patterns in classifier performance characteristics. With scattered coincidences being the dominant background component in clinical conditions, this suggests that while the method serves as an effective and geometry-agnostic replacement for traditional cut-based selection, meaningful further gains in image quality will require either larger input representations or dedicated treatment of the phantom-scattered component.

%%%%
\end{abstract}

\ifarxiv
  \maketitle
\fi

%\linenumbers
\section{Introduction}
\label{sec:Introduction}

\Ac{PET} is a widely used medical imaging modality that enables quantitative measurements of metabolic processes \textit{in vivo}.
One of the current trends in \ac{PET} scanner development is the introduction of \ac{LAFOV} systems with improved \ac{TOF} capabilities, which cover a considerably larger portion of the patient's body than conventional scanners.
The principal advantage of such systems is their significantly higher sensitivity, which can improve image resolution, enable dynamic whole-body imaging, shorten scan duration, or reduce the radiation dose delivered to the patients~\cite{surtiTotalBodyPET2020,slartlongaxialfield2021,karakatsanislongaxialfieldofview2026,esquinasSeeingMoreTreating2026}. 
Indeed, several such systems are already in clinical operation~\cite{prenosilperformancecharacteristicsbiograph2022a,badawiFirstHumanImaging2019,pantelpennpetexplorerhuman2020,yamagishiPerformanceCharacteristicsNewGeneration2023}, while additional concepts continue to be proposed and developed~\cite{vandenberghewalkthroughflatpanel2023,baranRealisticTotalbodyJPET2025a}.

However, the increased axial coverage of \ac{LAFOV} scanners operating in a fully 3D mode comes at a cost: a higher background level that degrades image quality.

In \ac{PET}, a radiopharmaceutical is administered to the patient. Depending on the tracer, it accumulates in regions characterised by a particular biological process -- for example, elevated glucose metabolism in the case of \ac{FDG}
The radioactive isotope undergoes $\beta^+$ decay, producing a positron that annihilates with an electron from the patient's body, resulting in the emission of a pair of nearly back-to-back 511\,keV photons. These photons are detected in coincidence by the scanner, with each detected event characterised by its deposited energy, interaction time, and spatial position within the detector.
Each pair defines an \ac{LOR}, and is passed to the image reconstruction algorithm.
The spatial density of reconstructed annihilation points forms an image that can be subsequently analysed by a physician, for example, to identify cancerous lesions.

Ideally, each coincidence corresponds to an unperturbed pair of photons originating from a single annihilation point (\textit{true} event). In practice, however, due to finite detector resolution and physical interaction processes, the coincidence set contains not only \textit{true} events but also a fraction of background pairs.
This background consists of three principal components.
First, accidental coincidences are formed from photons originating from different annihilation events, whose rates are of a combinatorial nature and increase approximately quadratically with the activity and axial length of the scanner. Second, phantom-scattered coincidences arise when one or both photons undergo Compton scattering within the patient's body before reaching the detector, thereby altering their direction and energy.  In the case of \ac{LAFOV} scanners, the contribution of multiply scattered photons becomes increasingly important. 
Finally, photons that undergo more than one interaction within the scanner detection system can be incorrectly assigned as a genuine photon pair, giving rise to so-called \textit{detector-scatter} events.
The occurrence and spatial distribution of these background coincidences depend on multiple factors, including scanner geometry, detector properties, and activity distribution within the patient.
For \ac{LAFOV} scanners, the acceptance of more oblique \ac{LOR}s increases the sensitivity but can lead to an increase in terms of the scatter and random fractions~\cite{daube-witherspoonTotalbodyPETNew2022}.

Conventional approaches to background suppression rely primarily on energy thresholds and geometrical constraints applied to the line of response. However, such methods are inherently limited in their ability to distinguish between signal and background events that share similar kinematic properties~\cite{Brasse2005}. Moreover, background reduction inevitably entails a partial loss of \textit{true} events, thereby reducing the overall system sensitivity. Consequently, background suppression in \ac{PET} represents a trade-off between maximising sensitivity and minimising background contamination.

The application of machine learning techniques to \ac{PET} data processing has grown considerably in recent years \cite{arabiPromiseArtificialIntelligence2021}, with \ac{ML} algorithms being used at various stages of the reconstruction pipeline, ranging from raw data filtering to enhancement of the reconstructed image \cite{readerDeepLearningPET2021}.

In the case of scattered and accidental coincidences, most of the recent works focus on corrections in image or sinogram space~\cite{laurent2023pet}. There are only a few existing approaches that target background reduction on an event-by-event basis at the raw coincidence level, prior to image reconstruction. Operating at this level is advantageous, as it allows individual photon properties to be taken into account, and the resulting method is not tied to a specific reconstruction method -- it can be naturally extended to multiphoton imaging, as well as to other imaging modalities such as Positron Emission Particle Tracking.
\cite{oliver_application_2013} considered the application of an \ac{ANN} for accidental coincidence reduction in small animal PET, a significantly simpler setting than clinical \ac{LAFOV} imaging.
As this work focused on accidental coincidences, neither scatter nor attenuation in the phantom was simulated. The authors did not consider factors such as the spatial uniformity of the classifier response. Instead, they used overall image quality metrics, such as the Contrast Recovery Coefficient (CRC) or the spillover ratio (SOR).
An interesting approach to recover, usually discarded, Inter-Crystal Scatter (ICS) triple coincidences was proposed by~\cite{michaud_sensitivity_2015}. An \ac{ANN} is used to classify the true photon pair within the three detected photons. The model does not classify phantom-scattered and accidental coincidences, and only standard accidental compensation is applied during image reconstruction.
While the authors verify their method using a Derenzo phantom, the model's performance is confounded by the reconstruction quality in this setting. As the spatial uniformity of the classifier response is not discussed, it is hard to judge if any dependence is observed.
A detailed study of various classifiers for background reduction from $^{176}$Lu radiation was presented by~\cite{wang_reduction_2020}.
In particular, the performance of \ac{XGBoost} and \ac{ANN} classifiers is compared; the focus is on reducing the $^{176}$Lu background, with classification of accidental and scattered coincidences left for further studies.
As in \cite{michaud_sensitivity_2015}, the authors validate their method using volumetric phantoms, including Derenzo and planar bars. Also, in this case, the spatial uniformity of the classifier response is not discussed.
In a recent study by~\cite{debuspetnetcoincidentparticle2024}, the use of Spiking Neural Networks was proposed as a replacement for traditional coincidence sorters. As this work focuses on what constitutes a coincidence, it does not address the problem of scattered coincidences.
Moreover, the authors consider only a centrally positioned point source, without accounting for possible dependence of the method's performance on the source position within the scanner.
While all of those works verify the performance of the classifier on out-of-distribution data, they are limited to training the machine learning model on a phantom that covers all or most of the scanner FOV and to application to other phantoms.
To our knowledge, no study has examined the impact of training data selection on the model performance for PET background reduction on an event-by-event basis.

In this contribution, we formulate coincidence classification as a supervised multi-class learning problem, applied event-by-event at the raw-data level, prior to image reconstruction, and evaluate its applicability to \ac{LAFOV} \ac{PET} background reduction. 
We compare two classifier families: \ac{BDT}s and \ac{ANN} networks trained on \ac{MC} simulations of the Siemens Biograph Vision Quadra scanner using two phantoms: the standard \ac{NEMA} phantom and the anthropomorphic \ac{XCAT} phantom. We perform a systematic study of feature selection, model generalisation across phantom geometries, and the impact of classifier hyperparameter selection.
In addition to standard classification metrics, we evaluate the classifier's impact on the reconstructed spatial distribution of the radio-tracer, which is crucial for image quality.
For this purpose, we propose an activity-agnostic method to visualise the model performance metrics, such as accuracy, as a three-dimensional map.
Such an approach enhances the model evaluation beyond reliance on global performance metrics. We show that a modest difference in a global metric, such as accuracy, can result from severe performance degradation in a particular image region, with no change in most of the volume.

Our findings establish the performance level of this approach with a minimal feature set and identify the conditions under which it can serve as a practical, geometry-agnostic replacement for traditional cut-based coincidence rejection. 
 
The rest of the article is structured into five sections. The Materials and Methods section~\ref{sec:Methods} begins with the formulation of the background reduction problem in terms of supervised classification with a description of the considered models. Next, we briefly define the evaluation metrics and discuss the input data generation and its characteristics. The section concludes with a discussion of model hyperparameter selection, followed by an overview of the feature selection.
The results section~\ref{sec:Results} presents the obtained classifier performance in terms of global performance metrics. 
A discussion of the impact of feature selection on the model performance and robustness to out-of-distribution data is provided in section~\ref{sec:Discussion}.
Finally, section~\ref{sec:Conclusions} provides the results summary and discusses the prospects and possible future extensions of the methods.

\section{Materials and Methods}
\label{sec:Methods}
\subsection{Machine Learning approach}
\label{subsec:methods_ml_models}
The problem of background reduction due to scatter and accidental coincidences can be formulated as a classification problem.
We consider a P-dimensional feature space $\mathcal{X} \in \mathbb{R}^{P}$. The vector $\vec{x} \in \mathcal{X}$ describes an event corresponding to a registered photon pair.  
The goal is to find a classifier function $f:\mathcal{X} \to \mathcal{Y}$ that maps the vector of features to a class type $\vec{y} \in \mathcal{Y}$.
In our case, the categorical output space $\mathcal{Y}$ consists of four classes corresponding to the physical coincidence types defined in section~\ref{subsec:methods_data}.
The optimal classifier $f^{*}$ is defined with respect to an objective function $O(f(\vec{x_i}), y_i)$, which typically consists of two parts: loss function $L(f(\vec{x_i}), y_i)$ and the regularization term $\Omega(f(\vec{x_i}))$. 
The loss function quantifies the quality of the prediction $f(\vec{x_i})$ for the given event described by the feature vector $\vec{x_i}$, compared to the expected class $y_i$.
The role of the regularisation term is to limit the model's complexity.
This helps to mitigate the problem of overfitting, the situation in which the model learns
perfectly the characteristics of the sample on which it was trained; however, it performs poorly when classifying events not seen before.
This optimisation task is solved by minimising the objective function, calculated based on 
the training sample composed of the $N$ events with the assigned labels: $T = [[\vec{x_1}, y_1], ...,[\vec{x_N}, y_N]]$.  Hence, the problem can be treated as an example of supervised learning.

Our classification models are based on two classes of machine learning algorithms: \ac{BDT} and \ac{ANN} networks.
Both model classes are well established, with their properties thoroughly studied. Moreover, compared to newer architectures, they are extremely fast.
Both properties make them good candidates for background reduction in a medical setting.

Our BDT classifiers are constructed based on 
libraries: \ac{XGBoost}~\cite{chenXGBoostScalableTree2016}, which is a current state-of-the-art \ac{BDT} implementation, and
\ac{AdaBoost}~\cite{freunddecisiontheoreticgeneralizationonline1997a}, one of the first robust \ac{BDT} algorithms.
In the following paragraph, the main ideas of decision tree-based classifiers and the boosting method are briefly discussed.
A decision tree -- \ac{CART} -- is a simple predictive model that can be used for classification or regression.
It consists of a chain of rules represented by nodes and branches.
Each node represents a test based on a single feature that splits the set of classified objects into subsets and assigns scores to the subsets.
The choice of a feature and the value at which the split is performed are determined by splitting rules.
The splitting process is repeated recursively until the subsets contain objects belonging to only one class, or until further splitting no longer improves the prediction score.
The terminal leaves are associated with the output classes. 
The classification based on the trained decision tree can be seen as a chain of 'yes-no' questions.
Since a single split corresponds to a linear division of the feature space along a given feature dimension, the final decision boundary, corresponding to the given tree model, divides the feature space into a hyperrectangle.

Decision trees are weak predictive models due to the high tendency to overfit.
Therefore, instead of using a single tree, the better approach is to use a set of decision trees and combine their predictions.
There are several methods of combining the set of trees.
Boosting is a general method of generating many simple classification rules (weak classifiers) and combining them into a single, highly accurate classifier~\cite{freunddecisiontheoreticgeneralizationonline1997a}.

This strategy is not limited to decision trees; rather, it is a general scheme and can be applied to other classification models, e.g., logistic regression or even deep neural networks.

The key idea  of boosting
consists of forming a
cumulative predictor by adding
subsequent weak classifiers sequentially, where each new classifier is 
trained to correct the mistakes of  its
predecessor. Formally, the prediction at boosting iteration $t$ is given by the additive model:

\begin{equation}
  \hat{y}_i^{(t)} = \sum_{k=1}^{t} f_k(\vec{x_i}) 
\end{equation}

where each $f_k$ denotes the $k-$th decision tree added to the ensemble and the sum runs over $t$ trees in the current iteration. The final ensemble prediction runs over all $T$ trees. To control model complexity and reduce overfitting, a regularisation term is added to the objective function for each tree $f$:
\begin{equation}
  \Omega(f) = \gamma L + \frac{1}{2} \lambda\sum_{j=1}^{L} w_j^2
\end{equation}
where the sum runs over all leaves $L$ of the tree, $ w_j$ is the score assigned to the $j-$th leaf, $\gamma$ is a penalty on the number of leaves that controls tree depth, and $\lambda$ is an L2 regularisation coefficient that penalises large leaf weights. 
Both $\gamma$ and $\lambda$ are hyperparameters, whose values are determined during the optimisation procedure described in  section~\ref{subsec:methods_hyperparameters}.

The trees are added greedily, one per iteration. At iteration $t$, the new tree
$f_t$ is chosen to minimise the regularised objective
\begin{equation}
  \mathcal{L}^{(t)} = \sum_i l\!\left(y_i,\, \hat{y}_i^{(t-1)} + f_t(\vec{x_i})\right)
  + \Omega(f_t),
\end{equation}
where $l$ is the per-sample loss (here the multiclass cross-entropy, i.e. the
\texttt{multi:softprob} objective). Since the loss cannot in
general be optimised directly over tree structures, \ac{XGBoost} approximates it
by a second-order Taylor expansion around the current prediction
$\hat{y}_i^{(t-1)}$, so each point contributes a gradient
$g_i = \partial_{\hat{y}} l$ and a second derivative $h_i = \partial^2_{\hat{y}} l$. For a fixed tree structure the objective is then
minimised analytically. Because the space of structures cannot be searched
exhaustively, each tree is grown by recursively splitting nodes, retaining at
every step the split that maximises the \emph{gain} defined as:

\begin{equation}\label{eq:xgb-gain}
    gain = \frac{1}{2}\left[\frac{G^2_L}{H_L+\lambda} + \frac{G^2_R}{H_R+\lambda} - \frac{(G_L+G_R)^2}{H_L+H_R+\lambda}\right]-\gamma
\end{equation}

with $G_{L(R)} = \sum_i g_i$ and $H_{L(R)} = \sum_i h_i$ the sums of the gradients
$g_i$ and second derivatives $h_i$ over the data points assigned to the left
(right) child after the split, and $\lambda$, $\gamma$ the regularisation terms.

\ac{AdaBoost} grows its \ac{CART} trees analogously, but
selects at each node the split maximising the Gini-impurity decrease, where impurity is defined in 
Eq.~(\ref{eq:gini-impurity}), and trains successive trees on reweighted samples so
that each focuses on its predecessor's errors.
The impurity decrease is the difference between a node's impurity and the Gini index --- a weighted sum of the impurity measures of the two child nodes. The impurity of a node $n$ is defined as:
\begin{equation}\label{eq:gini-impurity}
    \widehat{\Gamma}(n) = \sum^J_{j=1}\widehat{\phi}_j(n)\bigl(1-\widehat{\phi}_j(n)\bigr)
\end{equation}
where $\widehat{\phi}_j(n)$ denotes the frequency of class $j$ in node $n$.

For our \ac{ANN} model, we use the \ac{MLP} architecture. An \ac{MLP} is a foundational class of artificial neural networks. Structurally, an \ac{MLP} comprises at least three sequential layers of interconnected nodes: an input layer to receive the initial data, one or more hidden layers to extract underlying patterns, and an output layer to yield the final prediction.
Information propagates unidirectionally from input to output, with each node in the hidden and output layers serving as a computational unit with a non-linear activation function, such as a rectified linear unit (ReLU) or a sigmoid.
This non-linearity, combined with a hidden layer, enables the network to map high-dimensional feature spaces to target variables.
The network optimises its internal parameters through supervised learning with the backpropagation algorithm, which iteratively updates inter-node connection weights and node biases via gradient descent to minimise a predefined loss function.

Our \ac{MLP} implementation, built with the Keras library~\cite{chollet2015keras}, comprises a variable number $M$ of hidden layers, each containing a variable number $N$ of neurons. Both $N$ and $M$ are treated as hyperparameters of the model, and their selection is described in section~\ref{subsec:methods_hyperparameters}.
Each hidden layer utilises the ReLU activation function and is followed by a Dropout layer. The output layer is composed of several neurons equal to the number of considered coincidence classes, with Softmax as the activation function. The model is optimised using the Adam algorithm~\cite{kingma2014adam} with a sparse categorical cross-entropy loss function.

It should be noted that there exist additional, domain-specific requirements that a candidate classifier must fulfil.
For example, the classifier should be generic, in the sense that   
it cannot depend on the details of the phantom geometry.
Also, since the final objective is the patient's body image, the classification procedure cannot introduce any artefacts into the reconstructed image. 

\subsection{Hyperparameter selection}
\label{subsec:methods_hyperparameters}

The model hyperparameters are selected using Bayesian Optimisation~\cite{Brochu2010}.
It is a powerful optimisation technique for functions that are costly to evaluate, especially when derivatives are not available.
It is particularly well suited to hyperparameter optimisation of machine learning models.
The goal of this approach is to minimise the number of function evaluations by making informed decisions about which point in the hyperparameter space to evaluate next.
It is done by replacing the function with a surrogate model that is easy to evaluate and update with results of previous evaluations.
A common selection for the surrogate model is the Gaussian Process~\cite{Mockus1994}.
The selection of the next evaluation point is based on an acquisition function that balances exploration and exploitation.
Exploration is understood as probing areas with high variance in the surrogate model; hence, the probable gain is high but uncertain.
Exploitation involves probing regions where the optimised function is already known to be large.
A common selection for the acquisition function is the Expected Improvement~\cite{Jones1998}.
For the presented results, an open-source implementation of Bayesian Optimisation from the scikit-optimise package~\cite{headScikitoptimizeScikitoptimize2021} was used, with a Gaussian Process for the surrogate function and Expected Improvement as the acquisition function.
At each function evaluation by the procedure, a three-fold stratified cross-validation was performed, preserving the class ratios in each fold.
The mean accuracy from the three folds was used to update the surrogate model.
The best model was refit on the full training sample.

In Table~\ref{table:hyperparameters_ranges}, a list of considered hyperparameters for each model is presented along with the range of allowed values and the type of prior used for sampling the next point to evaluate, along with the results of the optimisation. An example of the obtained accuracy distribution as a function of parameter values is presented in Fig.~\ref{Fig:hyperparam_xgb_6f_example}. Distributions for all considered models and hyperparameters are provided in Appendix~\ref{appendix:hyperparamaters}.

\begin{figure}[!t]
\centering
  \subfloat[]{
    \includegraphics[width=0.49\linewidth]{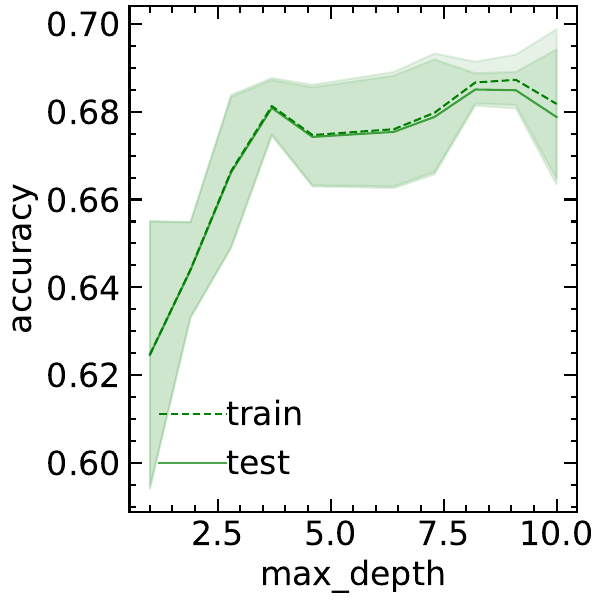}
  }
  \subfloat[]{
    \includegraphics[width=0.49\linewidth]{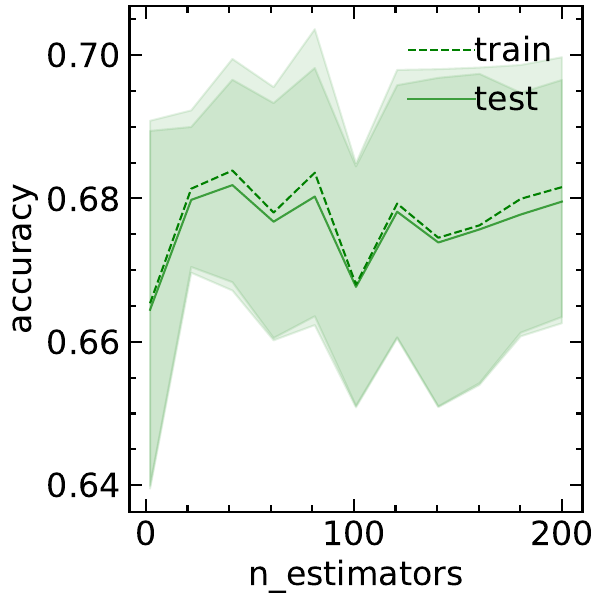}
  }
\caption{Example of observed dependence of the accuracy on model hyperparameters for the 6-feature \ac{XGBoost} model trained on the \ac{XCAT} sample.}
\label{Fig:hyperparam_xgb_6f_example}
\end{figure}

\subsection{Detector simulation}
\label{subsec:methods_simulations}

The studies were performed based on \ac{MC} simulations that modelled \ac{PET} measurements, carried out with the Siemens Biograph Vision Quadra, a \ac{LAFOV} PET/CT system~\cite{prenosilperformancecharacteristicsbiograph2022a}. The \ac{GATE} v9.4.1~\cite{sarrutAdvancedMonteCarlo2021,sarrutOpenGATEEcosystemMonte2022} simulation library, based on Geant v4.11.3.0~\cite{agostinelliGeant4SimulationToolkit2003}, has been used to model the data taking by the PET scanner. The interaction of particles with matter was simulated by using the em\_livermore\_polar physics list. Tracking of optical photons was switched off to reduce computation time. The simulation starts with isotropic emission of back-to-back photon pairs, with the initial energy set to $511$~keV.

The \ac{GATE} geometry of the scanner consists of 38 detection element arrays (rsectors) repeated in a ring with a radius of 420\,mm, replicated 4~times along the axial direction for a total axial field of view of 1060\,mm. Each rsector contains 8 detector modules arranged axially, each comprising $4 \times 2$ miniblocks of $5 \times 5$ lutetium oxyorthosilicate (LSO) crystals with individual dimensions of $3.2 \times 3.2 \times 20$\,mm$^3$, giving 243{,}200 crystals in total arranged in 320 crystal rings.

The front-end electronic response was modelled by the \ac{GATE} digitiser, which converts photon interactions in the scintillator crystals into deposited energy and detection time. Energy depositions within the same crystal were summed, and a readout grouping at the crystal level was applied. The energy resolution was parametrised as 9\% \ac{FWHM} at 511\,keV. Only single events with deposited energy within the 435--585\,keV window were accepted. Coincidences were formed from pairs of singles detected within a 4.7\,ns timing window, requiring a minimum sector difference of~1. The timing resolution was applied in post-processing by smearing the detection times with a Gaussian of $\sigma_t = 93$\,ps, corresponding to a \ac{CRT} of 219\,ps. Additionally, the position of each photon interaction was discretised to the centre of the crystal in which the energy was deposited.

\subsection{Phantoms}
\label{subsec:methods_phantoms}

\begin{figure}[htb]
\centering {
  \subfloat[]{
    \label{Fig:phantoms_reco_xcat}
    \includegraphics[width=\linewidth]{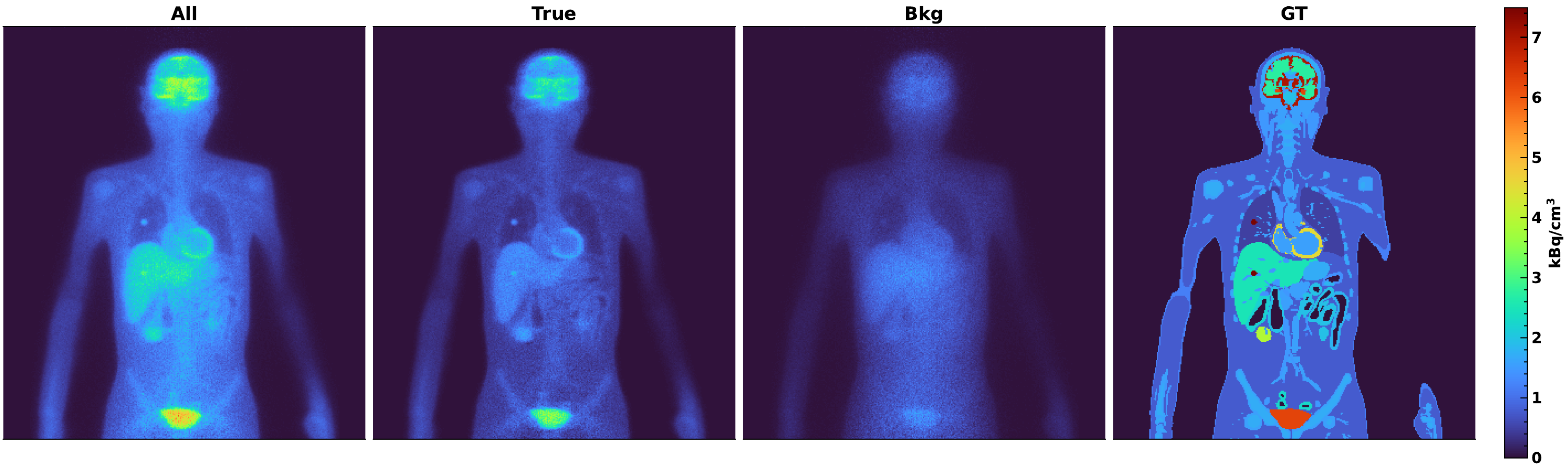}
  }\\
  \subfloat[]{
    \label{Fig:phantoms_reco_nema}
    \includegraphics[width=\linewidth]{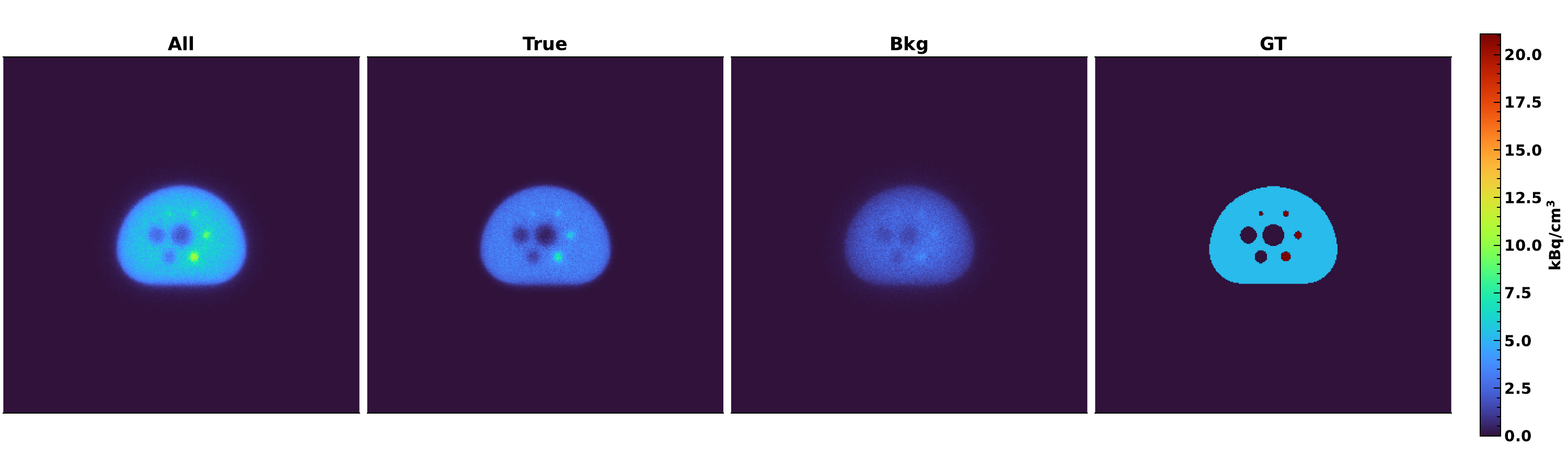}
  }
}
\caption{\label{Fig:phantoms_reco}Reconstructed activity for {\it ALL}, {\it TRUE}, background ({\it PHT}, {\it DET} and {\it ACC}) coincidences as well as the \ac{GT} activity map for \ac{XCAT} (a) and NEMA IEC (b) phantoms.}
\end{figure}

Two experimental setups were chosen for the simulations: \ac{NEMA}~\cite{NEMA:2018} and \ac{XCAT}~\cite{segars4DXCATPhantom2010a} phantoms. 
The 22~cm long \ac{NEMA} phantom is built out of four high-activity (denoted as {\it hot}) and two low-activity (denoted as {\it cold}) spheres. The phantom is positioned isocentrically within the scanner. Hot spheres of 10~mm, 13~mm, 17~mm and 22~mm diameters and cold spheres of 28~mm and 37~mm diameters were simulated. The ratio between the hot spheres and the background was set to 4:1. The total activity of 50 MBq was simulated, and the acquisition time was set to 100 seconds. 

The activity of the male \ac{XCAT} phantom was prepared ~\cite{zincirkeser2007standardized, silva2015simulated} to mimic the $^{18}$F-FDG distribution within the human body. Additionally, two hot spheres (diameter = 1.2 cm) positioned in the lung and liver were incorporated into the phantom simulations. The contrast between the hot region and the background activity was set to 16:1 and 3:1 for the lungs and liver, respectively. The phantom's overall activity was 50 MBq, and the acquisition time was set to 1000 seconds. The \ac{XCAT} phantom activity map is depicted in Fig.~\ref{Fig:phantoms_reco_xcat}, while the \ac{NEMA} phantom activity map is depicted in Fig.~\ref{Fig:phantoms_reco_nema}.

For each dataset, we extract randomly two subsamples: a training sample of $2.4\cdot10^6$ coincidences and an independent testing sample of $0.6\cdot10^6$ coincidences. As described in section~\ref{subsec:methods_hyperparameters}, the models are trained using a three-fold stratified cross-validation; hence, each cross-validation iteration is trained on $1.6\cdot10^6$ coincidences. Remaining coincidences are used for image reconstruction and generation of quality maps as described in sections~\ref{subsec:methods_image_reconstruction}~and~\ref{subsec:methods_quality_maps}. In addition, the best model is evaluated on out-of-distribution data via cross-inference: the \ac{NEMA} trained model is evaluated on the \ac{XCAT} data sample, and the \ac{XCAT} model on the \ac{NEMA} dataset.

\subsection{Data representation}
\label{subsec:methods_data}
We now define the output space $\mathcal{Y}$ and the input feature space $\mathcal{X}$ in terms of the physical properties of the detected coincidences.

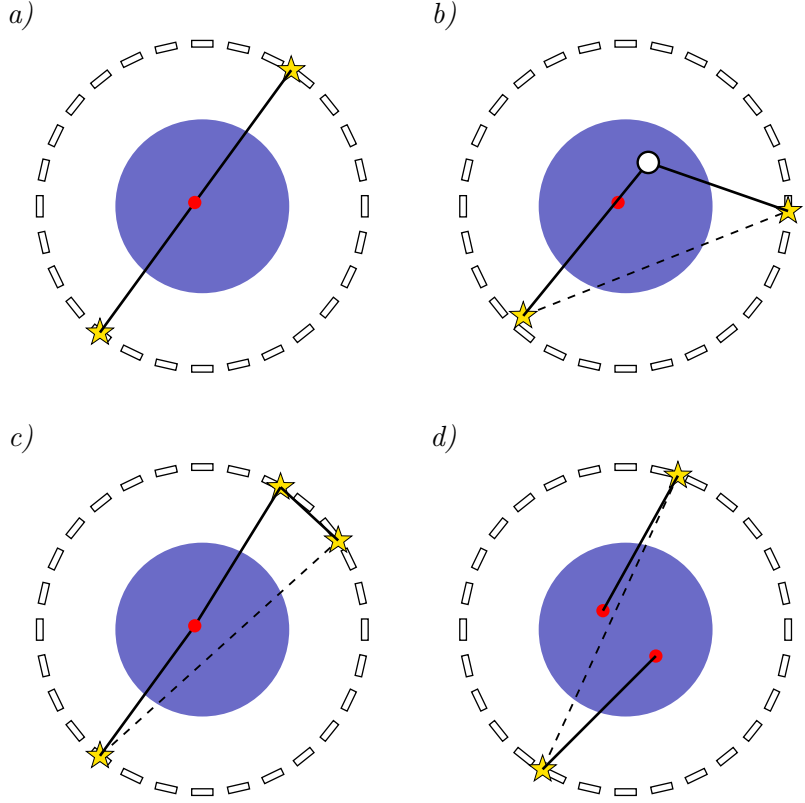
\begin{figure}[htb]
\centering
\begin{tikzpicture}
 
\def\dx{5.6}
\def\dy{5.6}
\def\Rring{2.15}
\def\Ndet{28}
\def\Rbody{1.15}
 
% ============================================================
% a) TRUE  --  top-left
% LOR is a straight line passing exactly through the red dot (-0.1, 0.05).
% Star positions computed so they are collinear with the red dot at ring radius.
% ============================================================
\begin{scope}[shift={(0,\dy)}]
  \node[anchor=south east, font=\itshape] at (-\Rring+0.05, \Rring+0.05) {a)};
  \scannerring{0}{0}{\Rring}{\Ndet}
  \node[body, minimum size=2*\Rbody cm] at (0,0) {};
  % Draw line first, then red dot on top so it is visible
  \drawstar{-1.354}{-1.670}{0.19}
  \drawstar{1.176}{1.800}{0.19}
  \draw[black, line width=1pt] (-1.354,-1.670) -- (1.176,1.800);
  \node[emissionpt] at (-0.1, 0.05) {};
\end{scope}
 
% ============================================================
% b) PHT  --  top-right
% Scatter circle is well inside the body. The line is deflected at the
% scatter point: segment 1 from bottom-left star to scatter point,
% segment 2 from scatter point to top-right star (different direction).
% ============================================================
\begin{scope}[shift={(\dx,\dy)}]
  \node[anchor=south east, font=\itshape] at (-\Rring+0.05, \Rring+0.05) {b)};
  \scannerring{0}{0}{\Rring}{\Ndet}
  \node[body, minimum size=2*\Rbody cm] at (0,0) {};
  \node[emissionpt] at (-0.1, 0.05) {};
  % Scatter point well inside the body (radius ~0.65, body radius 1.15)
  \coordinate (scat) at (0.30, 0.58);
  % Dotted line connecting both stars (expected LOR)
  \draw[black, line width=0.7pt, dashed] (-1.354,-1.45) -- (2.149,-0.063);
  % Segment 1: from bottom-left star to scatter point (original direction)
  \drawstar{-1.354}{-1.45}{0.19}
  % Segment 2: deflected at 110 deg -- exit star computed for 110 deg between semilines
  \drawstar{2.149}{-0.063}{0.19}
  \draw[black, line width=1pt] (-1.354,-1.45) -- (0.30,0.58);
  \draw[black, line width=1pt] (0.30,0.58) -- (2.149,-0.063);
  % Void circle drawn on top of the kink
  \node[scatterpt] at (scat) {};
\end{scope}
 
% ============================================================
% c) DET  --  bottom-left
% Both kink stars lie exactly on the scanner ring (radius 2.15).
% kink1 = (1.035, 1.884), kink2 = (1.798, 1.179) -- both normalised to Rring.
% ============================================================
\begin{scope}[shift={(0,0)}]
  \node[anchor=south east, font=\itshape] at (-\Rring+0.05, \Rring+0.05) {c)};
  \scannerring{0}{0}{\Rring}{\Ndet}
  \node[body, minimum size=2*\Rbody cm] at (0,0) {};
  % Dotted line: registered LOR connecting BL star to final detection star (kink2)
  \draw[black, line width=0.7pt, dashed] (-1.354,-1.670) -- (1.798,1.179);
  % Straight photon: BL star to red dot (no star at top end)
  \drawstar{-1.354}{-1.670}{0.19}
  \draw[black, line width=1pt] (-1.354,-1.670) -- (-0.1,0.05);
  % Kinked photon: red dot -> kink1 -> kink2, stars at kink1 and kink2
  \coordinate (kink1) at (1.035, 1.884);
  \coordinate (kink2) at (1.798, 1.179);
  \drawstar{1.035}{1.884}{0.19}
  \drawstar{1.798}{1.179}{0.19}
  \draw[black, line width=1pt] (-0.1,0.05) -- (kink1);
  \draw[black, line width=1pt] (kink1) -- (kink2);
  % Red dot drawn last, so it sits on top of the straight line
  \node[emissionpt] at (-0.1, 0.05) {};
\end{scope}
 
% ============================================================
% d) ACC  --  bottom-right
% Two separate emission vertices. Each emits one photon as a semi-line
% going outward to the scanner ring -- no connection between the two rays.
% ============================================================
\begin{scope}[shift={(\dx,0)}]
  \node[anchor=south east, font=\itshape] at (-\Rring+0.05, \Rring+0.05) {d)};
  \scannerring{0}{0}{\Rring}{\Ndet}
  \node[body, minimum size=2*\Rbody cm] at (0,0) {};
  % Dotted line: expected LOR connecting both registered stars
  \draw[black, line width=0.7pt, dashed] (-1.098,-1.848) -- (0.692,2.036);
  % Vertex 1 at (-0.30, 0.25) -- photon goes top-right to (0.692, 2.036)
  \node[emissionpt] at (-0.30, 0.25) {};
  \drawstar{0.692}{2.036}{0.19}
  \draw[black, line width=1pt] (-0.30, 0.25) -- (0.692, 2.036);
  % Vertex 2 at (0.40, -0.35) -- photon goes bottom-left to (-1.098, -1.848)
  \node[emissionpt] at (0.40, -0.35) {};
  \drawstar{-1.098}{-1.848}{0.19}
  \draw[black, line width=1pt] (0.40,-0.35) -- (-1.098,-1.848);
\end{scope}
 
\end{tikzpicture}
\caption{Classes of the coincidences $Y$. The blue circle represents the patient's body, and the small
rectangles positioned circularly denote the scanner. The red point is the emission vertex. The yellow
stars represent the registered photon position. The dotted line represents the incorrectly reconstructed \ac{LOR}. a) \textit{TRUE}: genuine photon pair. b) \textit{PHT}: one of the
photons is scattered in the patient's body (void circle) before reaching the scanner. c) \textit{DET}: one of
the photons scatters twice in the scanner. d) \textit{ACC}: two photons originating from different emission
vertices are classified as a coincidence.}
\label{fig:eventTypes}
\end{figure}

Each event is assigned a class label from the following four categories, corresponding to the coincidence types illustrated in Fig.~\ref{fig:eventTypes}:
 
\begin{itemize}
\item genuine, undisturbed photon pairs (denoted as {\it TRUE}), 
\item photon pairs with at least one scatter in the patient's body (denoted as {\it PHT} ),
\item photon pairs with at least one additional interaction in the detector system (denoted as {\it DET}),
\item accidental coincidences from uncorrelated annihilation events (denoted as {\it ACC}). 
\end{itemize}
 
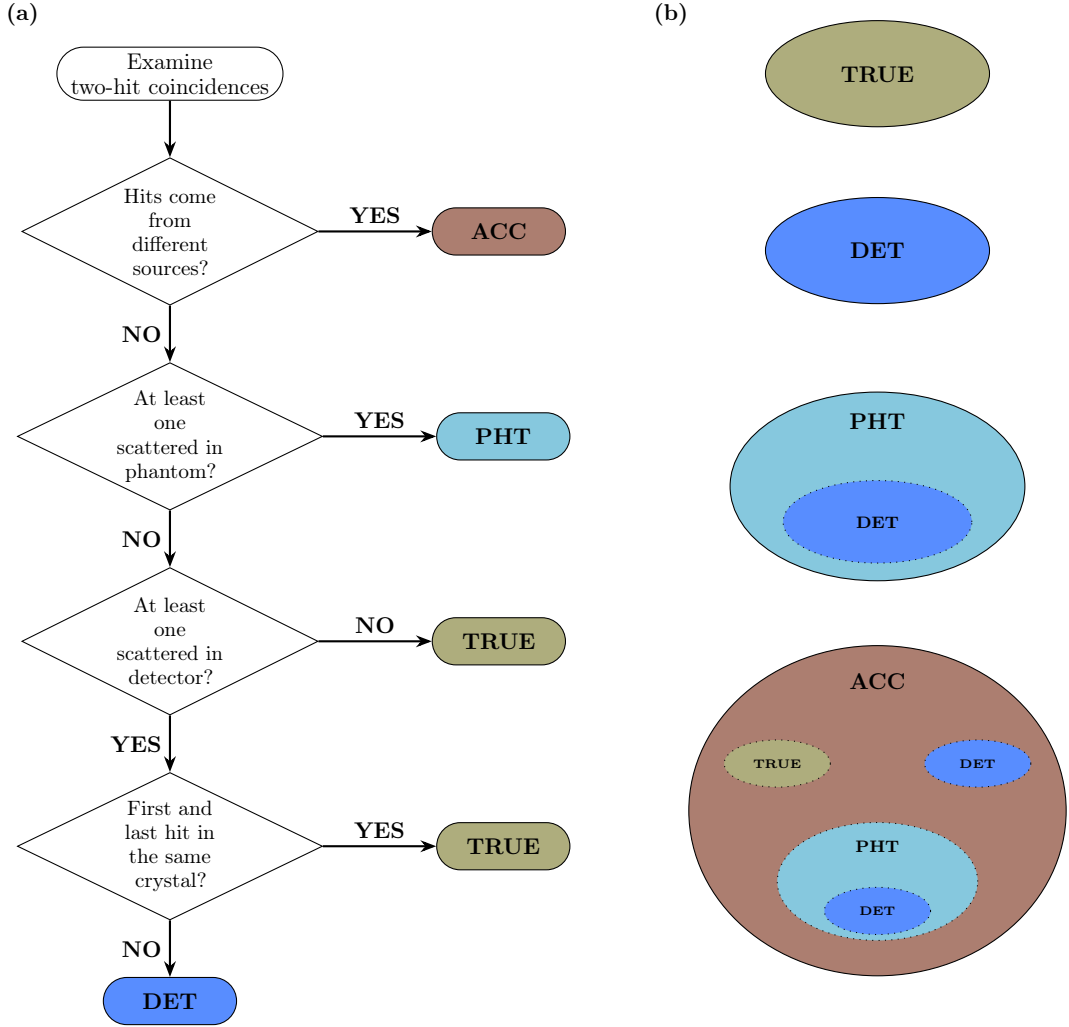
\begin{figure}[htb]
\centering
\begin{tikzpicture}[node distance=0.75cm, scale=0.78, every node/.style={scale=0.78},
  decision/.append style={inner sep=0pt, font=\footnotesize\linespread{0.8}\selectfont,
    minimum width=2.5cm, minimum height=2.5cm, text width=1.8cm, align=center},
  startstop/.append style={inner sep=2pt, font=\small\linespread{0.85}\selectfont},
  result/.append style={inner sep=2pt}]

  % Color definitions
  \definecolor{colTRUE}{HTML}{AEAD7D}
  \definecolor{colDET}{HTML}{588DFF}
  \definecolor{colPHT}{HTML}{86C8DE}
  \definecolor{colACC}{HTML}{AC7E70}
  \definecolor{colSTART}{HTML}{C2B8A8}

  % ============== LEFT PANEL: FLOWCHART ==============

  \node[startstop, fill=white] (start) {Examine\\two-hit coincidences};

  \node[decision, below=0.75cm of start] (d1)
    {Hits come from different sources?};

  \node[result, fill=colACC,
        right=1.5cm of d1] (random) {ACC};

  \node[decision, below=0.75cm of d1] (d2)
    {At least one scattered in phantom?};

  \node[result, fill=colPHT,
        right=1.5cm of d2] (phscat) {PHT};

  \node[decision, below=0.75cm of d2] (d3)
    {At least one scattered in detector?};

  \node[result, fill=colTRUE,
        right=1.5cm of d3] (true1) {TRUE};

  \node[decision, below=0.75cm of d3] (d4)
    {First and last hit in the same crystal?};

  \node[result, fill=colTRUE,
        right=1.5cm of d4] (true2) {TRUE};

  \node[result, fill=colDET,
        below=0.75cm of d4] (detscat) {DET};

  % --- Arrows ---
  \draw[arrow] (start) -- (d1);
  \draw[arrow] (d1) -- node[above, font=\small\bfseries] {YES} (random);
  \draw[arrow] (d1) -- node[left, font=\small\bfseries] {NO} (d2);
  \draw[arrow] (d2) -- node[above, font=\small\bfseries] {YES} (phscat);
  \draw[arrow] (d2) -- node[left, font=\small\bfseries] {NO} (d3);
  \draw[arrow] (d3) -- node[above, font=\small\bfseries] {NO} (true1);
  \draw[arrow] (d3) -- node[left, font=\small\bfseries] {YES} (d4);
  \draw[arrow] (d4) -- node[above, font=\small\bfseries] {YES} (true2);
  \draw[arrow] (d4) -- node[left, font=\small\bfseries] {NO} (detscat);

  % Label (a)
  \node[font=\bfseries] at (-2.5, 1.0) {(a)};

  % ============== RIGHT PANEL: CLASS ADMIXTURES ==============

  \begin{scope}[xshift=10.5cm, yshift=-1.0cm]

  % Label (b)
  \node[font=\bfseries] at (-2.0, 2.0) {(b)};

  % TRUE (standalone)
  \node[draw, ellipse, fill=colTRUE,
        minimum width=3.8cm, minimum height=1.8cm,
        font=\bfseries\small]
    at (1.5, 1.0) {TRUE};

  % DET (standalone)
  \node[draw, ellipse, fill=colDET,
        minimum width=3.8cm, minimum height=1.8cm,
        font=\bfseries\small]
    at (1.5, -2.0) {DET};

  % PHT > DET
  \node[draw, ellipse, fill=colPHT,
        minimum width=5.0cm, minimum height=3.2cm]
    at (1.5, -6.0) {};
  \node[font=\bfseries\small] at (1.5, -4.9) {PHT};
  \node[draw, dotted, ellipse, fill=colDET,
        minimum width=3.2cm, minimum height=1.4cm,
        font=\bfseries\scriptsize]
    at (1.5, -6.6) {DET};

  % ACC > {TRUE, DET, PHT > DET}
  \node[draw, ellipse, fill=colACC,
        minimum width=6.4cm, minimum height=5.6cm]
    at (1.5, -11.5) {};
  \node[font=\bfseries\small] at (1.5, -9.3) {ACC};

  \node[draw, dotted, ellipse, fill=colTRUE,
        minimum width=1.8cm, minimum height=0.8cm,
        font=\bfseries\tiny]
    at (-0.2, -10.7) {TRUE};

  \node[draw, dotted, ellipse, fill=colDET,
        minimum width=1.8cm, minimum height=0.8cm,
        font=\bfseries\tiny]
    at (3.2, -10.7) {DET};

  \node[draw, dotted, ellipse, fill=colPHT,
        minimum width=3.4cm, minimum height=2.0cm]
    at (1.5, -12.7) {};
  \node[font=\bfseries\scriptsize] at (1.5, -12.1) {PHT};
  \node[draw, dotted, ellipse, fill=colDET,
        minimum width=1.8cm, minimum height=0.8cm,
        font=\bfseries\tiny]
    at (1.5, -13.2) {DET};

  \end{scope}

\end{tikzpicture}

\caption{(a) Scheme of the algorithm assigning class labels to the coincidences based on the information
from \ac{MC} simulations. (b) Representation of the four main event classes together with the potential subsets of events that originate from the assumed classification hierarchy.
}
\label{fig:labeling}
\end{figure}

Class labels are assigned based on the \ac{MC} truth information available from the simulation (see Fig.~\ref{fig:labeling}(a)).
The decomposition of $\mathcal{Y}$ into these four classes is not unique, but it is governed by the order of the selection rules. Because the rules are evaluated sequentially (Fig.~\ref{fig:labeling}), an event labelled as {\it ACC} may also involve a photon that scattered in the phantom or detector; had the labelling order been different, such an event would have been assigned to {\it PHT} or {\it DET} instead. Independently of the labelling order, some events within one class are indistinguishable from another class in the observable feature space -- for instance, an accidental coincidence composed of two unscattered photons has the same measured characteristics as a {\it TRUE} event. Both types of admixtures are illustrated in Fig.~\ref{fig:labeling}: the dotted subsets represent events that, due to labelling priority, share characteristics with the enclosed class.

The resulting class distributions for both simulated phantoms are presented in Table~\ref{table:class_fractions}.

\begin{table}[h!]
\centering
\begin{tabular}{| c | c | c |} 
 \hline
 Class & NEMA IEC & \ac{XCAT} \\ 
 \hline\hline
 {\it TRUE} & 46.8 \% & 49.0 \% \\
 {\it PHT}  & 25.2 \% & 24.5 \% \\
 {\it DET}  &  9.8 \% & 10.2 \% \\
 {\it ACC}  & 18.2 \% & 16.3 \% \\
 \hline
\end{tabular}
\caption{Class fractions for the NEMA IEC and \ac{XCAT} simulation samples after geometry cuts.}
\label{table:class_fractions}
\end{table}

For selected tests, we consider a 3-class scenario (3C), in which we merge the {\it TRUE} and {\it DET} classes. We observe that both classes exhibit similar photon characteristics, making it difficult for the classifier to distinguish between them.

\begin{figure}[htb]
\centering {
  \subfloat[]{
    \includegraphics[width=0.49\linewidth]{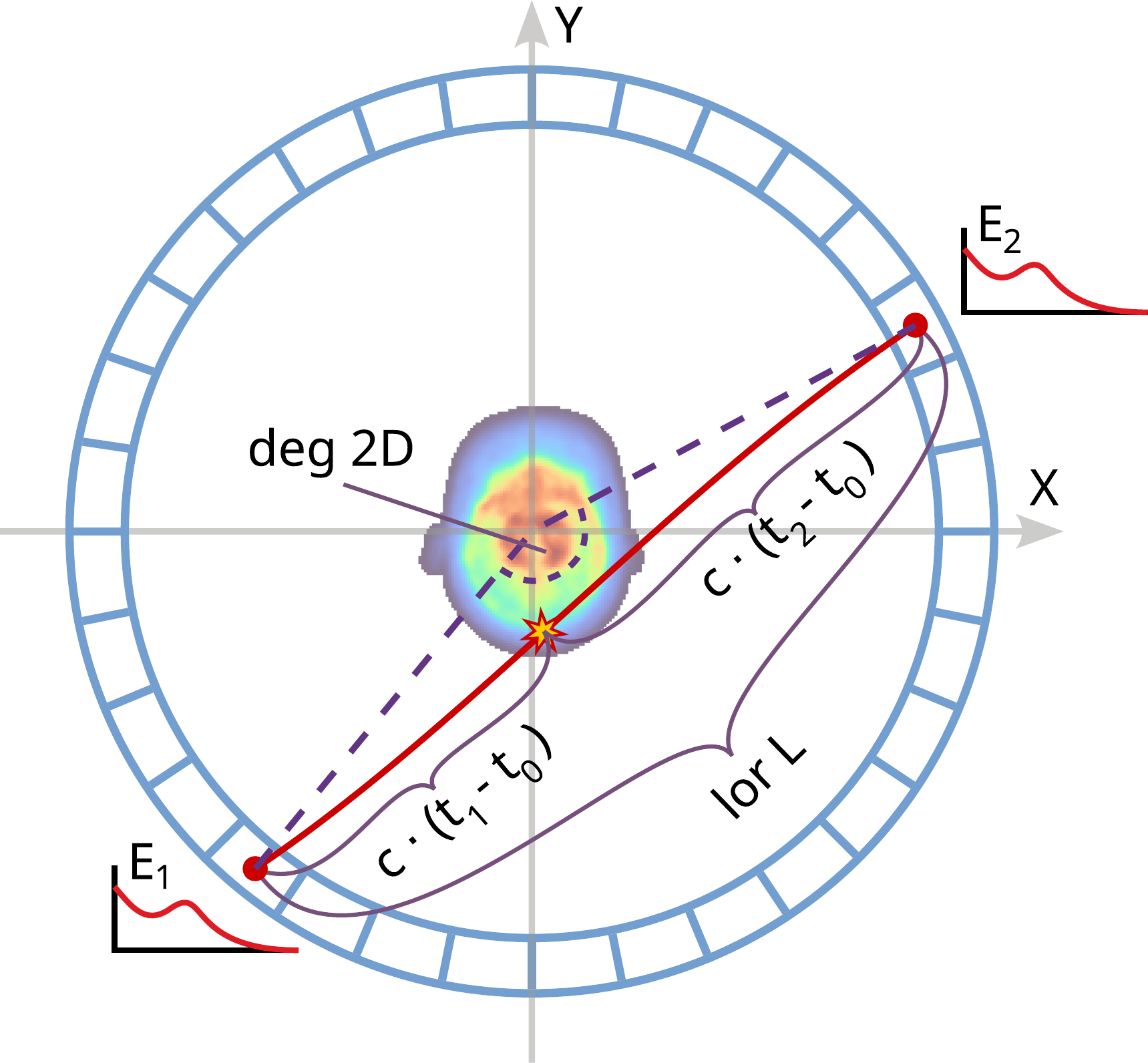}
  }
  \subfloat[]{
    \includegraphics[width=0.49\linewidth]{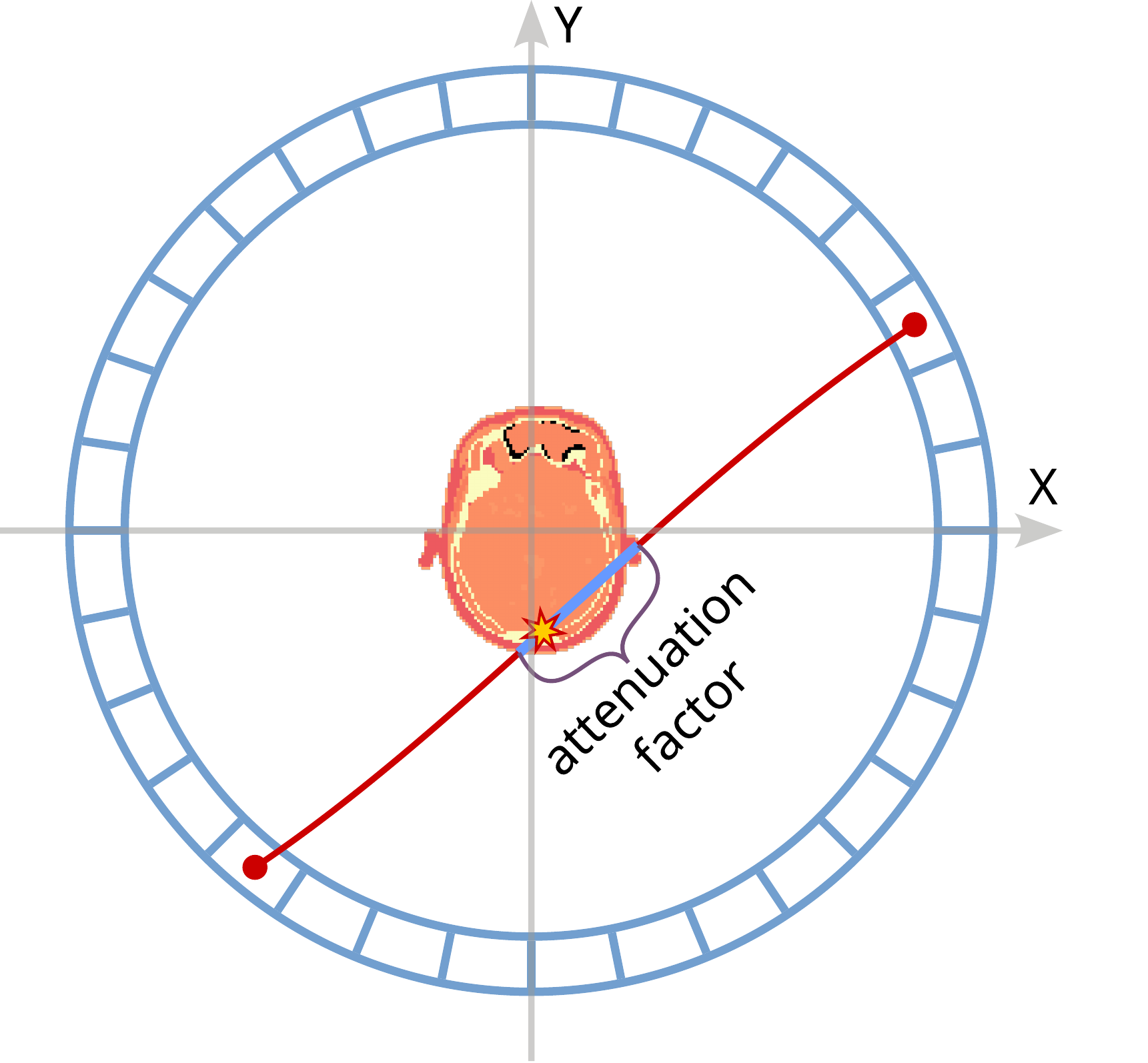}
  }
}
\caption{\label{Fig:features_diagram} Graphical representation of observables used to create the final feature set. Panel (a) presents the $deg2D$ opening angle, $lorL$, photon detection $t_1$, $t_2$ and emission $t_0$ times together with photon deposited energies $E_1$ and $E_2$. Panel (b) presents the way the attenuation factor is extracted for a given coincidence.}
\end{figure}

Each {\it event} is described by ten scalar features corresponding to the following parameters: the reconstructed spatial positions $x_1$, $y_1$, $z_1$, and  $x_2$, $y_2$, $z_2$,  of the two registered photons, their deposited energies $E_1, E_2$, and their registration times $t_1, t_2$.
These raw quantities are not directly suitable as classification features for two reasons.
First, certain raw measurements carry no discriminating information on their own. For example, the absolute photon registration times $t_1$ and $t_2$ are meaningless without a common time reference.  
Moreover, coincidences are typically formed by sorting photons by time or energy, which introduces an artificial skewness in the feature distributions of the two photons.
Since we always consider each coincidence as an unordered photon pair without distinguishing their identities, the constructed features should be symmetric with respect to the operation of photon swap: $x^p(1,2) = x^p(2,1)$, where 1 and 2 denote the first and second photon in the pair. 
The second important point to consider is that the distributions of the x-, y-, and z-positions of the registered hit directly encode the detector and phantom geometries; thus, this choice is not ideal for maintaining the model's generality.
We therefore construct the following set of derived features (Fig.~\ref{Fig:features_diagram}), all of which satisfy the photon swap symmetry condition by construction:
\begin{itemize}
 \item absolute time difference between the registered hits:
  \begin{equation}
   dt = |t_1 - t_2|;
  \end{equation}
 \item absolute value of the deposited energy difference:
  \begin{equation}
   eDiff=|E_{1} - E_{2}|;
  \end{equation}
 \item sum of deposited energies:
  \begin{equation}
   eSum = E_1 + E_2;
  \end{equation}
 \item \ac{AF} computed by ray-tracing through a map of linear \SI{511}{keV} attenuation coefficients $\mu$ - which is normally obtained from the \ac{CT} scan
  \begin{equation}
    AF = \exp \left( -\int_{0}^{lorL} \mu(x) dx \right).
  \end{equation}
\end{itemize}

In addition, two topology-dependent features are defined: 
\begin{itemize}
 \item 2D opening angle with respect to the geometrical centre of the detector
  \begin{equation}
   deg2D = \arccos{\left(\frac{x_1 \times x_2 + y_1 \times y_2}{\sqrt{{x_1}^2 +{y_1}^2} \times \sqrt{{x_2}^2 +{y_2}^2}}\right)};
  \end{equation}
 \item length of the line of response spanned between two registered hits
  \begin{equation}
   lorL = \sqrt{(x_1 -x_2)^2 +(y_1 -y_2)^2 +(z_1 -z_2)^2 }.
  \end{equation}
\end{itemize}

We consider two scenarios: 4-features (4F) and 6-features (6F).
In the former, only the first group of previously described variables is used, and the final feature space becomes four-dimensional $\mathcal{X} \in \mathbb{R}^{4}$.
For the latter, all six variables are considered, leading to a six-dimensional $\mathcal{X} \in \mathbb{R}^{6}$ feature space.
Such a distinction is made to assess the impact of topological features on the model's generality.
In both scenarios, the output space $\mathcal{Y}$ is defined as a categorical one with four possible outcomes $Y = [TRUE, PHT, DET, ACC]$, as defined above.  

All features are preprocessed using the min-max normalisation:
\begin{equation}
   x^{(p)}_{i} \to 
   \frac{x_i^{(p)} - \min\limits_{1 \leq j \leq N} x_j^{(p)}}
    {\max\limits_{1 \leq j \leq N} x_j^{(p)} - \min\limits_{1 \leq j \leq N} x_j^{(p)}},
   \qquad 
   i \in [1,...,N], p \in [1,...,P]  
\end{equation}
The minimum and the maximum values for each ($p$) of $P$ features are computed over the training sample containing $N$ events.
The normalisation parameters are determined during the training procedure and subsequently reused for validation and inference.

\subsection{PET image reconstruction}
\label{subsec:methods_image_reconstruction}
A direct \ac{TOF} positioning algorithm is used for image reconstruction. Each coincidence defines a \ac{LOR} from two photon detection points. 
Based on the time difference $dt$ between the photon registrations, the most likely annihilation position is estimated along the \ac{LOR} and the event
is deposited into the corresponding voxel of a 3D grid with 2.5 mm isotropic spacing ($424 \times 280 \times 280$ voxels).
The intensity in each voxel is thus defined by the number of coincidences whose estimated emission point falls within that voxel volume. 

The reconstructed image is corrected for attenuation and detector sensitivity using a combined attenuation--sensitivity map derived using the CASToR~\cite{merlincastorgenericdata2018} reconstruction package from a dedicated \ac{GATE} simulation. The corrected image is then normalised to the assumed phantom activity.

This reconstruction approach is chosen deliberately for two reasons. First, it is parameter-free, unlike iterative algorithms such as \ac{MLEM}, it introduces no tunable hyperparameters that could interact with or obscure the classifier's effect on image quality. Second, and crucially for the present study, the direct \ac{TOF} positioning preserves an explicit one-to-one correspondence between each coincidence event and its deposited voxel. This property allows mapping the spatial distribution of classifier metrics, such as accuracy, TPR, and PPV, directly onto the reconstructed image volume, as described in section~\ref{subsec:methods_quality_maps}. In contrast, iterative algorithms redistribute event contributions across the image through repeated forward and backwards projection steps, which would destroy event-to-event traceability and introduce additional systematic effects into the spatial quality maps.

The reconstructed images for both phantoms are presented in Fig.~\ref{Fig:phantoms_reco}, showing separate reconstructions for {\it ALL}, {\it TRUE}, and background ({\it PHT}, {\it DET} and {\it ACC}) coincidence classes, with class association based on the true label of each event from the \ac{MC} simulation.

\subsection{Evaluation metrics}
\label{subsec:methods_eval_metrics}
We evaluated the performance of the classifiers based on several metrics commonly used in ML, expressing the dependence between the genuine, positive (P) and spurious, false (F) classification:\\
True Positive Rate (sensitivity)~\footnote{There are several names for the same metric depending on the scientific community. The True Positive Rate is also known as sensitivity, recall or efficiency, while the Positive Predictive Value is known as purity, or precision.} 
\begin{equation}
TPR = \frac{TP}{TP + FN};
\end{equation}
Positive Predictive Value (precision):
\begin{equation}
PPV = \frac{TP}{TP + FP};
\end{equation}
accuracy
\begin{equation}
accuracy = \frac{TP + TN}{TP + TN + FP + FN};
\end{equation}
F1 Score
\begin{equation}
F1 = \frac{2TP}{2TP + FP + FN};
\end{equation}
\ac{MCC}
\begin{equation}
MCC = \frac{TP \cdot TN - FP \cdot FN}{\sqrt{(TP + FP)(TP + FN)(TN + FP)(TN + FN)}};
\end{equation}
with $T$ and $F$ denoting, respectively, true and false classification into P and N classes.

As the model performance is not uniform between coincidence classes, we defined a confusion matrix for each model.
A visual representation of model prediction correctness, with rows corresponding to true event class and columns to model predictions, while matrix elements are a count of events for a given pair of true and predicted label (section~\ref{subsec:results_model_performance}).

We verify two modes of true event selection based on the model response.
For the {\it threshold} mode, we select an optimal working point by considering the TPR as a function of PPV -- the \ac{ROC} curve.
As a reference point, we selected the model response threshold corresponding to a TPR (sensitivity) of 95\% and treated all events that passed this threshold as the {\it TRUE} class.
When the model performance was evaluated on a different dataset, both the data normalisation and the classifier working points were kept as defined during training to simulate application to real-world data, where selection of a new working point is not possible.
For the {\it top-1} mode, the model response is the class with the largest model-assigned probability for a given event. Based on this, we select the events with the {\it TRUE} class being the top-1.

In addition, the \ac{MSE} metric was used.
The \ac{MSE} between two images $I_{1}$ and $I_{2}$ is defined as:
\begin{equation}
MSE[I_1, I_2] = \frac{1}{N}\sum_{k=1}^{N} (I_1[k] -I_2[k])^2, 
\end{equation}
where N is the total number of voxels.

In our studies, the ToF-reconstructed, {\it TRUE} class image is used as a ground-truth reference, while the predicted image is a ToF reconstruction of ML-selected coincidences.
Both the reference and ML-predicted images are first corrected using a combined scanner sensitivity and attenuation map and normalised to the simulated activity. Then, a median filter with kernel $2^3$ is applied to both images.
The \ac{MSE} calculation is restricted to voxels within the phantom volume by a mask derived from the attenuation map.

\subsection{Spatial quality maps}
\label{subsec:methods_quality_maps}

The model performance can vary spatially depending on the local environment of the emission point, e.g. activity, tissue type, tissue density, and position within the scanner \ac{FOV}.
Commonly used global metrics, such as accuracy or \ac{MCC}, fail to capture localised degradations of model performance, while such spatial non-uniformity can be detrimental for medical diagnosis.
We therefore propose complementing global metrics with 3D spatial maps to identify degraded regions.
First, we divide all coincidences into TP, FP, TN, and FN subgroups based on the model's response.
Although both binary and multi-class definitions of this division are applicable, we adopt a binary distinction between the signal and the background for two reasons.
First and foremost, the determination of the {\it TRUE} class is the ultimate goal of the classification.
Second, the binary formulation allows direct comparison between two proposed classification modes: \textit{top-1} and \textit{threshold}.

Each coincidence subgroup is then reconstructed independently as described in section~\ref{subsec:methods_image_reconstruction}, providing a per-voxel estimate of TP, FP, TN and FN.
Correction for attenuation and sensitivity is not required, as it cancels out when calculating the metrics defined in section \ref{subsec:methods_eval_metrics}. Finally, by calculating in each voxel three metrics: accuracy, TPR and PPV, we obtain 3D maps of model quality metrics. Resulting maps for the \ac{XGBoost} model are presented in Fig.~\ref{Fig:metric_maps_robustness_xcat}~and~\ref{Fig:metric_maps_robustness_nema}.

\subsection{Feature analysis}
\label{subsec:methods_feature_analysis}

To gain insight into the final model's characteristics, it is useful to determine the feature importance metric, a relative measure of the feature's usefulness for the classification task. Two approaches were considered: one for the \ac{BDT} classifiers, where such metrics are naturally available, and the permutation feature importance for all considered models. In the case of the \ac{XGBoost}, the feature importance is commonly an average {\it gain} (Eq.~\ref{eq:xgb-gain}) across all leaves and all trees in the ensemble where the feature was used.
For the \ac{AdaBoost}, the {\it gini importance} is defined as the sum of node impurity decrease weighted by the fraction of samples reaching the node averaged over all trees in the ensemble. The impurity decrease (Eq.~\ref{eq:gini-impurity}) is the difference between a node’s impurity and the Gini index - a weighted sum of the impurity measures of the two child nodes.

No such intrinsic metric exists for artificial neural networks; therefore, a good solution, albeit computationally intensive, is to determine the permutation feature importance~\cite{Breiman2001}.
It is defined as a reduction of a model score when applied to data in which one feature is randomly shuffled.
By removing the relationship between the feature and the target, the model's performance will reflect how crucial the feature's information is.

\section{Results}
\label{sec:Results}

\subsection{Feature analysis}
\label{subsec:results_feature_analysis}

\begin{figure}[htb]
\centering {
  \subfloat[]{
    \includegraphics[width=0.33\linewidth]{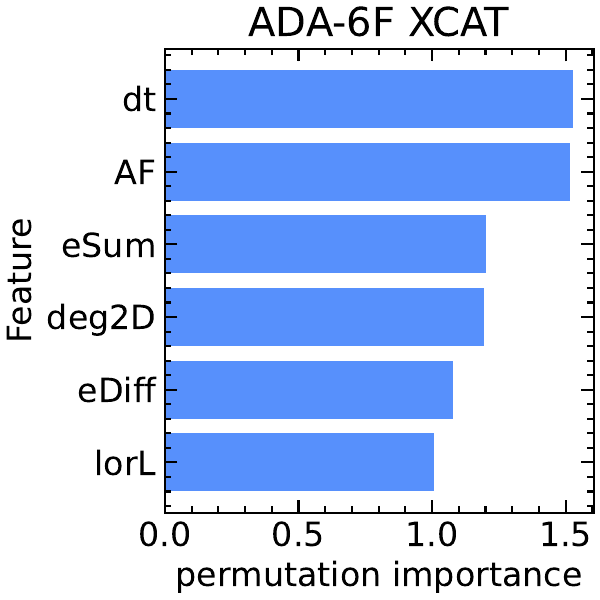}
  }
  \subfloat[]{
    \includegraphics[width=0.33\linewidth]{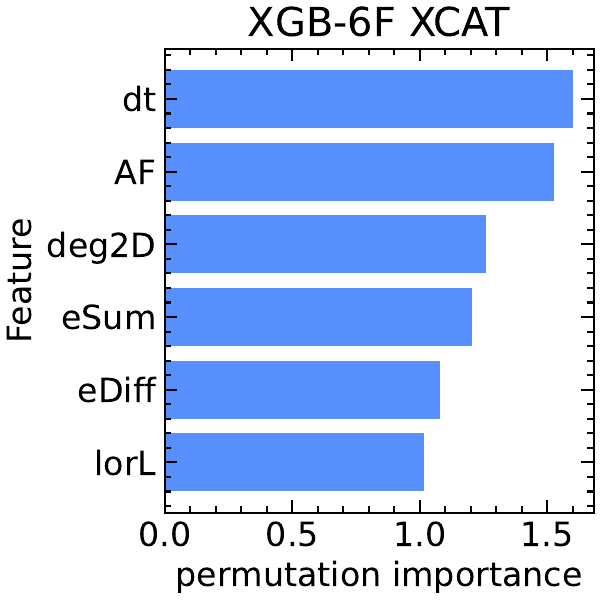}
  }
  \subfloat[]{
    \includegraphics[width=0.33\linewidth]{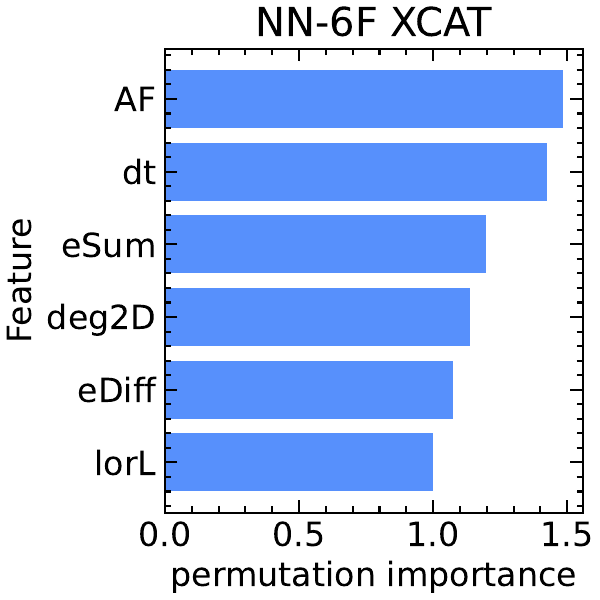}
  }
}
\caption{\label{fig:permutation_feature_importance_example}Permutation feature importance for the three considered models for the \ac{XCAT} sample.}
\end{figure}

We present the permutation importance of the six features for the three considered model classes, obtained for the \ac{XCAT} dataset. The two features with the highest score for all three models are $dt$ and $AF$, as presented in Fig.~\ref{fig:permutation_feature_importance_example}. The first two features have similar importance values and form a distinct group, with $eSum$ and $deg2D$ features forming a second one. We observe that the ordering within closely valued groups is model-dependent.

Full results of the feature importance analysis are presented in Appendix~\ref{appendix:feature_importance}. We observe that for the \ac{NEMA} dataset, the $deg2D$ gains importance over $AF$ due to the simplified geometry. It is worth noting that, unlike permutation importance, the \ac{XGBoost} gain metric identifies $eSum$ as the most important feature across both datasets. While essentially free to compute, the \ac{XGBoost} gain includes the regularisation terms, which are harder to interpret; therefore, a model-agnostic metric such as permutation importance is preferable.

\subsection{Models performance}
\label{subsec:results_model_performance}

\begin{figure}[htb]
\centering {
  \subfloat[]{
    \includegraphics[width=0.49\linewidth]{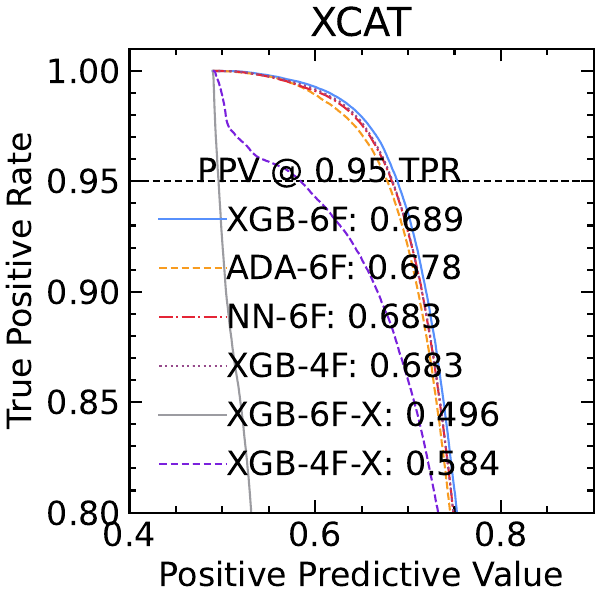}
  }
  \subfloat[]{
    \includegraphics[width=0.49\linewidth]{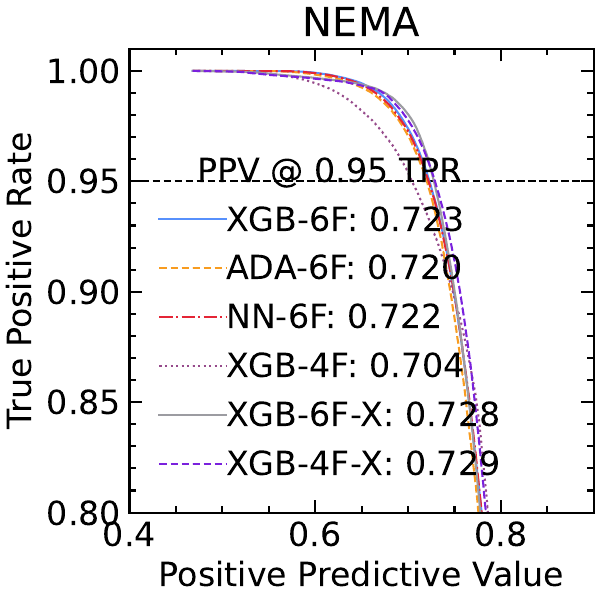}
  }
}
\caption{\label{fig:purity_vs_sensitivity}TPR as a function of PPV for all considered model configurations, evaluated on the \ac{XCAT} (a) and NEMA IEC (b) samples.}
\end{figure}

\begin{figure}[htb]
\centering {
  \subfloat[]{
    \includegraphics[width=0.49\linewidth]{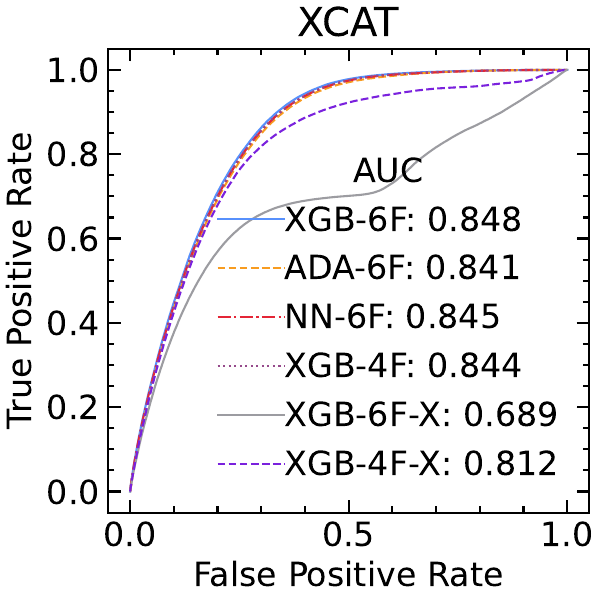}
  }
  \subfloat[]{
    \includegraphics[width=0.49\linewidth]{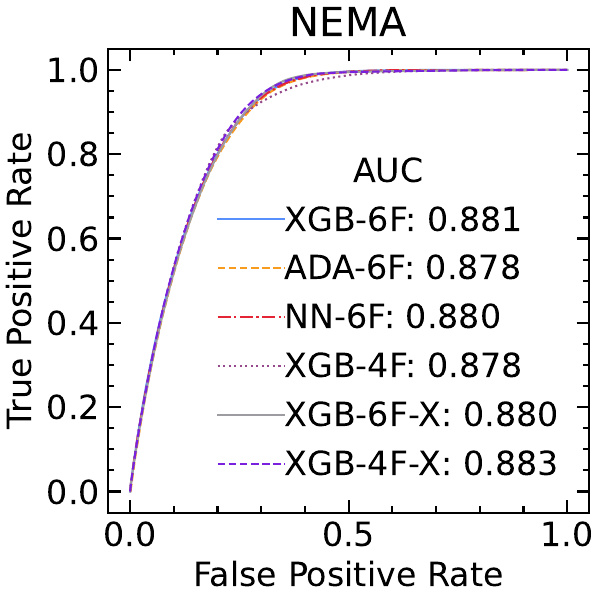}
  }
}
\caption{\label{fig:roc}\ac{ROC} curves for all considered model configurations, evaluated on the \ac{XCAT} (a) and NEMA IEC (b) samples.}
\end{figure}

We present results of the coincidence classification for two datasets: \ac{NEMA} and \ac{XCAT}. For each dataset, a total of six model configurations are considered. First, we evaluate the three model types trained on the full six-feature set: \ac{XGBoost} (XGB-6F), \ac{AdaBoost} (ADA-6F) and \ac{MLP} (NN-6F). For the best architecture (\ac{XGBoost}), we provide three additional tests: a variant trained on the four geometry-agnostic features (XGB-4F) and two cross-inference results in which the model is evaluated on an out-of-distribution dataset (XGB-6F-X and XGB-4F-X).

The performance metrics for the best variants of each architecture are provided in Table~\ref{table:best_model_metrics}. In addition, we present the fraction of {\it TRUE} coincidences for the baseline geometric cut, together with the model binary accuracy, for direct comparison.

\begin{table}[h!]
\centering
\begin{tabular}{| c | c c c | c c c |} 
 \hline
                      & \multicolumn{3}{c|}{NEMA IEC} & \multicolumn{3}{c|}{\ac{XCAT}} \\ 
%\hline
                      & XGB   & ADA   & MLP   & XGB   & ADA   & MLP \\ 
 \hline\hline
 Baseline             &       & 0.468 &       &       & 0.490 & \\
 Binary Accuracy      & 0.806 & 0.804 & 0.806 & 0.765 & 0.754 & 0.760 \\
 Multi-class Accuracy & 0.738 & 0.737 & 0.734 & 0.691 & 0.689 & 0.685 \\
 MCC                  & 0.60  & 0.60  & 0.60  & 0.51  & 0.51  & 0.50 \\
 F1                   & 0.71  & 0.71  & 0.70  & 0.66  & 0.65  & 0.65 \\
 \hline
\end{tabular}
\caption{Performance metrics for the best variant - 6-features and 4 classes - of each architecture: \ac{XGBoost}, \ac{AdaBoost} and \ac{MLP}. The three left columns correspond to the \ac{NEMA} phantom while the three right columns correspond to the \ac{XCAT} one. The baseline corresponds to the fraction of {\it TRUE} coincidences after applying the geometry cuts; binary accuracy is computed by treating all classes besides the {\it TRUE} as background. Multi-class accuracy, MCC and F1 metrics are described in the section~\ref{subsec:methods_eval_metrics}.}
\label{table:best_model_metrics}
\end{table}

The TPR dependence on PPV for all configurations is presented in Fig.~\ref{fig:purity_vs_sensitivity}, with numeric PPV values for binary classification with a model response threshold set to obtain a TPR equal to 0.95. As described in section~\ref{subsec:methods_eval_metrics}, it is the working point of the {\it threshold} selection mode. A clear degradation in performance is observed for the \ac{NEMA} trained model when evaluated on out-of-distribution data. In Fig.~\ref{fig:roc}, the receiver operating characteristic curve is presented for the considered configurations together with quoted values of the associated Area Under Curve (AUC) metric.

\begin{figure}[htb]
\centering {
  \subfloat[]{
    \includegraphics[width=0.49\linewidth]{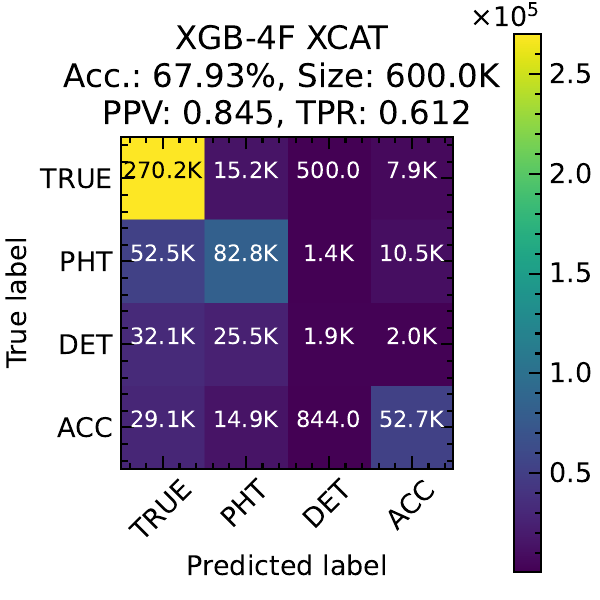}
  }
  \subfloat[]{
    \includegraphics[width=0.49\linewidth]{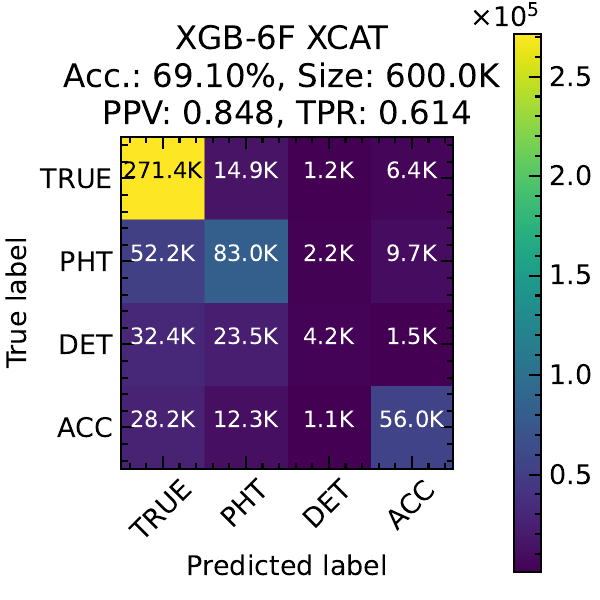}
  }
}
\caption{\label{fig:confusion_matrices_xcat}Confusion matrices for the best \ac{XGBoost} models for the \ac{XCAT} sample are presented. The left column (a) corresponds to a model trained with 4 features, while the right column (b) presents results for a 6-feature model variant.}
\end{figure}

\begin{figure}[htb]
\centering {
  \subfloat[]{
    \includegraphics[width=0.49\linewidth]{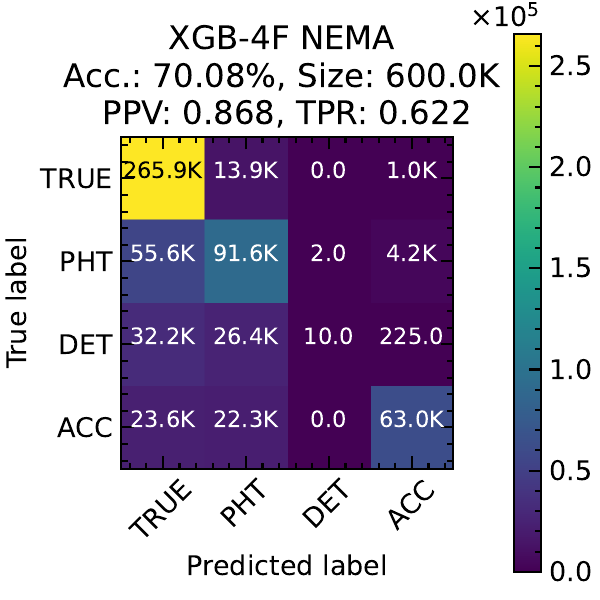}
  }
  \subfloat[]{
    \includegraphics[width=0.49\linewidth]{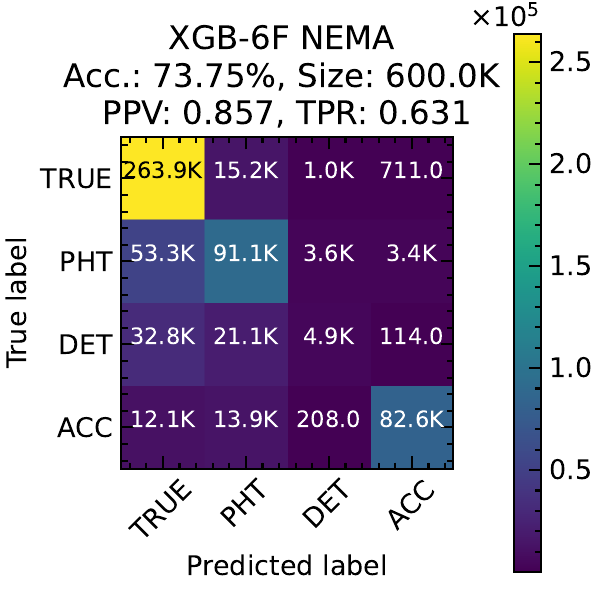}
  }
}
\caption{\label{fig:confusion_matrices_nema}Confusion matrices for the best \ac{XGBoost} models for the \ac{NEMA} sample are presented. The left column (a) corresponds to a model trained with 4 features, while the right column (b) presents results for a 6-feature model variant.}
\end{figure}

For the best architecture (\ac{XGBoost}), we present in Figs.~\ref{fig:confusion_matrices_xcat}~and~\ref{fig:confusion_matrices_nema} confusion matrices evaluated on the test subsamples for the two datasets: \ac{NEMA} and \ac{XCAT}. In addition, the values of the multi-class Accuracy, PPV, and TPR metrics are quoted. It's evident that among the studied configurations, models struggle most to correctly classify the {\it DET} class. This is not unexpected, as in events where the photon after Compton scattering deposits energy in a neighbouring crystal, the features are hardly perturbed compared to the {\it TRUE} class. On the other hand, such coincidences do not introduce large image smearing. Another class for which models struggle is the {\it PHT} class, where a substantial part of the coincidences is wrongly classified as {\it TRUE}.

\begin{figure}[htb]
\centerline {
\includegraphics[width=\linewidth]{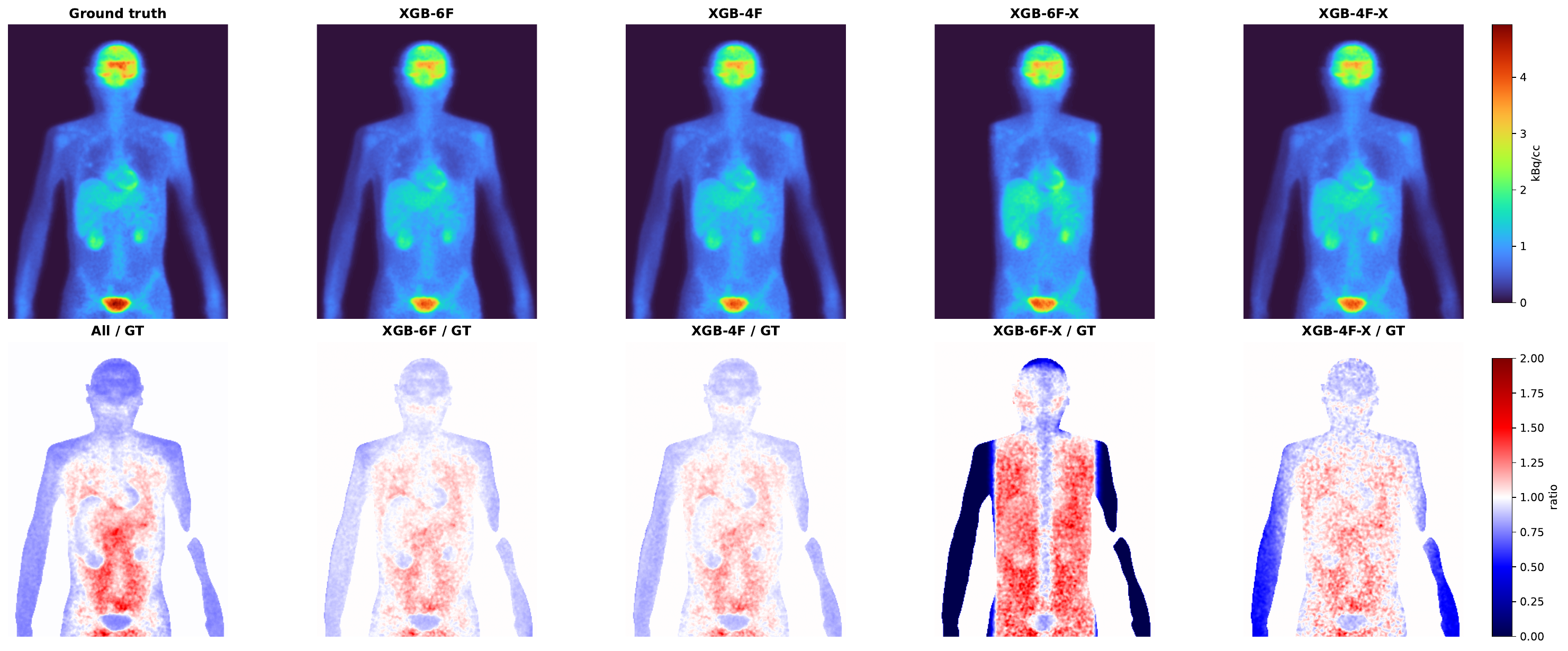}}
\caption{\label{fig:reconstruction_mse_xcat}Top row: sagittal slices of activity normalized reconstructed \ac{XCAT} images for {\it TRUE} coincidences (ground truth) and \ac{XGBoost} selected coincidences in four variants: two trained on the \ac{XCAT} sample with six (XGB-6F) and four features (XGB-4F), and two trained on the \ac{NEMA} sample also with six (XGB-6F-X) and four (XGB-4F-X) features. Bottom row: per-voxel ratio maps masked to the phantom volume - GT/All shows the {\it TRUE} class fraction, other columns follow the same convention as the top row and present the ratio of classifier-selected coincidences to {\it TRUE} ones.}
\end{figure}

A visual representation of the image quality for the best model on the \ac{XCAT} sample is presented in Fig.~\ref{fig:reconstruction_mse_xcat}. The XGB-6F selected coincidences were reconstructed with corrections for the scanner sensitivity and attenuation as described in section~\ref{subsec:methods_image_reconstruction}. The image is compared to the reconstruction of the {\it TRUE} class coincidences. A clear improvement is seen compared to the uncorrected image.

\subsection{Robustness test of the models}
\label{subsec:results_robustness_test}
In the proposed method, models are trained on a single phantom.
This raises a question about the generality of the resulting models, particularly with respect to phantom geometry.
To verify this, we devised a cross-inference test where a model trained on one phantom is evaluated on a dataset from another phantom.
For this work, models trained on the NEMA IEC phantom were evaluated on the \ac{XCAT} sample and models trained on the \ac{XCAT} were evaluated using the NEMA IEC phantom.
The \ac{XCAT} phantom is not only more than 7 times longer than NEMA IEC, but its geometry is also much more complex.
Therefore, with this test, we can evaluate both the impact of the phantom's position within the scanner and of different phantom shapes or orientations.

% Wyniki
We observe that 6-feature models exhibit worse generalisation performance than the 4-feature case, where topological features are not considered.
For example, the \ac{XGBoost} model trained on the \ac{NEMA} data sample and evaluated on the \ac{XCAT} data sample achieved accuracies of $0.561$ for the 6-feature variant and $0.647$ for the 4-feature one.
Comparing this result to the performance of models trained on the \ac{XCAT} sample, we observe a loss of accuracy of $0.130$ and $0.044$ for the 6- and 4-feature variants, respectively.
The accuracy gap of the 4-feature \ac{XGBoost} model to the 6-feature one, trained on the \ac{NEMA} sample, when evaluated on the same phantom is $0.037$. This shows that it exhibits good generalisability as its performance profile is stable independent of the data sample. More interestingly, the performance gap narrows for models trained on the \ac{XCAT} data sample to only $0.012$; for a phantom with more diverse geometry, the topological features do not allow for simple yet robust geometry cuts. Once again, the gap is stable when considering cross-inference on the \ac{NEMA} phantom.

\begin{figure}[htb]
\centering {
  \subfloat[]{
    \includegraphics[width=0.49\linewidth]{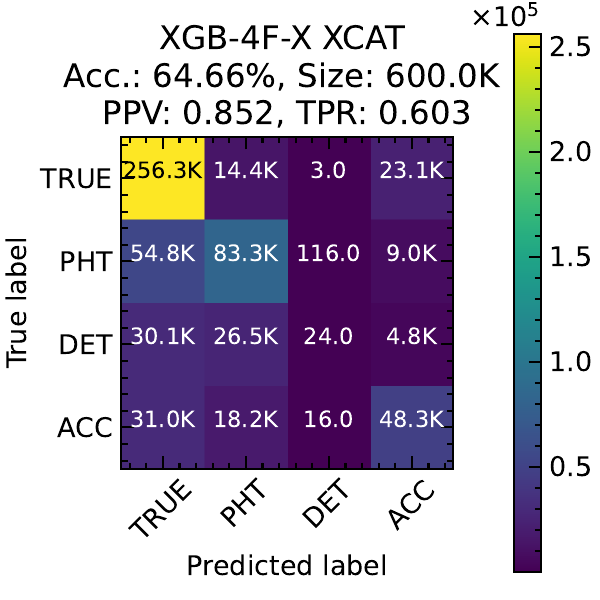}
  }
  \subfloat[]{
    \includegraphics[width=0.49\linewidth]{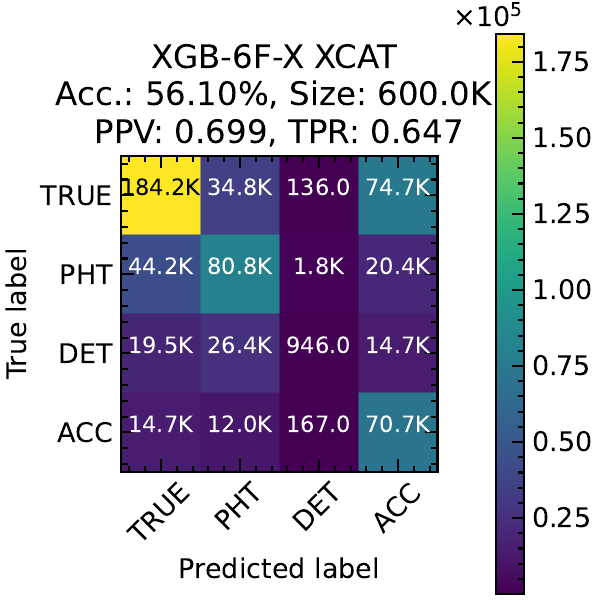}
  }\\
  \subfloat[]{
    \includegraphics[width=0.49\linewidth]{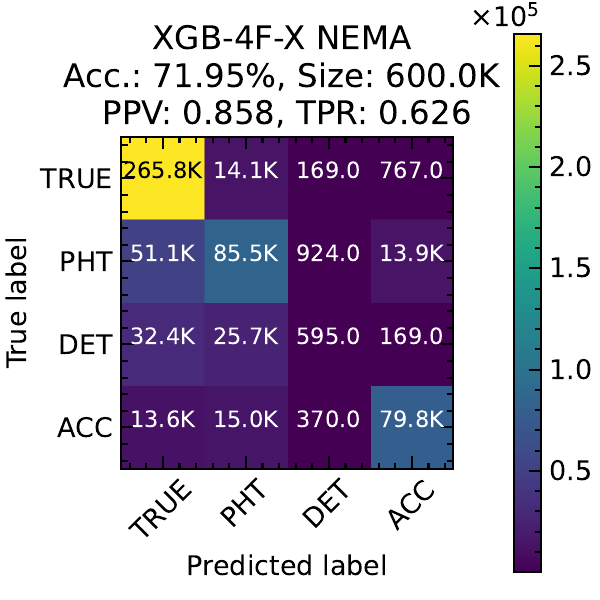}
  }
  \subfloat[]{
    \includegraphics[width=0.49\linewidth]{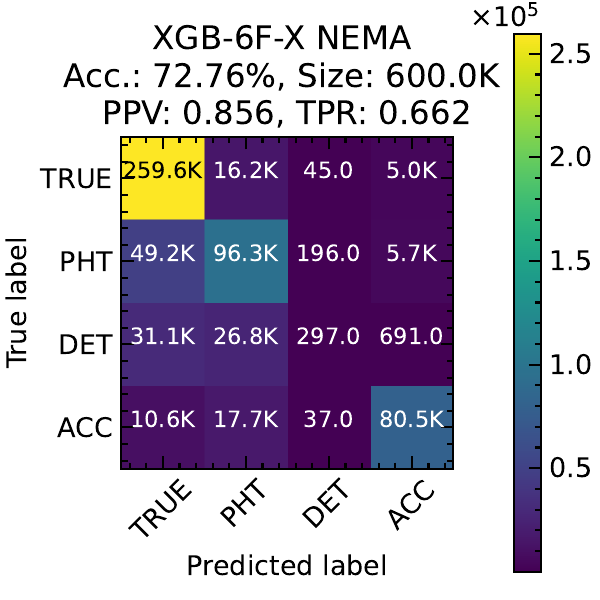}
  }
}
\caption{\label{fig:robustness_confusion_matrices}Confusion matrices for the robustness test. Results for \ac{XGBoost} models are presented. In the first row (figures a) and b)) models trained on the NEMA IEC sample are evaluated on \ac{XCAT} sample. For the bottom row (figures c) and d)) models trained on \ac{XCAT} sample are evaluated on the NEMA IEC sample.
The left column (figures a) and c)) corresponds to models trained with 4-features while the right column (figures b) and d)) presents results for 6-feature model variants.}
\end{figure}

The results of the robustness test are summarised in Table~\ref{table:robustness} and in Fig.~\ref{fig:robustness_confusion_matrices}.
Evaluating classifier performance with a multi-class version of the accuracy metric allows for an insight into how well it discerns between the background classes in addition to the {\it TRUE} one.
Based on the obtained confusion matrices, the 4-feature variants do not discern as well as the 6-feature ones between {\it PHT} and {\it DET} classes.

\begin{table}[h!]
\centering
\renewcommand{\arraystretch}{1.5} % Adds uniform vertical padding to all cells
\begin{tabular}{| c | >{\centering\arraybackslash}p{0.13\linewidth} | >{\centering\arraybackslash}p{0.15\linewidth} | >{\centering\arraybackslash}p{0.15\linewidth} | >{\centering\arraybackslash}p{0.15\linewidth} | >{\centering\arraybackslash}p{0.15\linewidth} |} 
 \hline
 & & \multicolumn{4}{c|}{\ac{XGBoost} model variant} \\
 \cline{3-6}
 sample & metric               & 4F NEMA        & 4F \ac{XCAT}         & 6F NEMA              & 6F \ac{XCAT} \\
 \hline\hline
 \multirow{3}{2em}{\centering \rotatebox[origin=c]{90}{\ac{XCAT}}} 
 & accuracy & 0.647 (-0.044)  & \underline{0.679} (-0.012)        & 0.561 (-0.130)           & \textbf{0.691} (0.0) \\
 \cline{2-6}
 & F1                   & 0.60 (-0.05)   & \underline{0.64} (-0.01)         & 0.54 (-0.11)         & \textbf{0.65} (0.0) \\
 \cline{2-6}
 & MCC                  & 0.44 (-0.07)   & \underline{0.49} (-0.02)         & 0.36 (-0.15)         & \textbf{0.51} (0.0) \\
 \hline
 \multirow{3}{2em}{\centering \rotatebox[origin=c]{90}{NEMA IEC}} 
 & accuracy & 0.701 (-0.037)  & 0.720 (-0.018)         & \textbf{0.738} (0.0) & \underline{0.728} (-0.010) \\
 \cline{2-6}
 & F1                   & 0.66 (-0.05)   & 0.68 (-0.03)         & \textbf{0.71} (0.0)  & \underline{0.69} (-0.02) \\
 \cline{2-6}
 & MCC                  & 0.54 (-0.06)   & 0.57 (-0.03)         & \textbf{0.60} (0.0)  & \underline{0.58} (-0.02) \\
 \hline
\end{tabular}
\caption{Results of the robustness tests for \ac{XGBoost} models. The first column identifies the evaluation data sample. The columns correspond to results for four models: 4F NEMA - a model trained on \ac{NEMA} data sample with 4-features, 4F \ac{XCAT} - a model trained on \ac{XCAT} data sample with 4-features, 6F NEMA - a model trained on \ac{NEMA} data sample with 6-features and 6F \ac{XCAT} - a model trained on \ac{XCAT} data sample with 6-features. The quoted values correspond to the accuracy, F1 and MCC metrics defined for multi-class classification. In parentheses, the metric reduction relative to the best result for the given data sample is provided. Best results are provided in bold while the second best are underlined.}
\label{table:robustness}
\end{table}

\subsection{Spatial quality maps}
\label{subsec:results_quality_maps}

\begin{figure}[htbp]
\centering {
  \subfloat[]{
    \includegraphics[width=\linewidth]{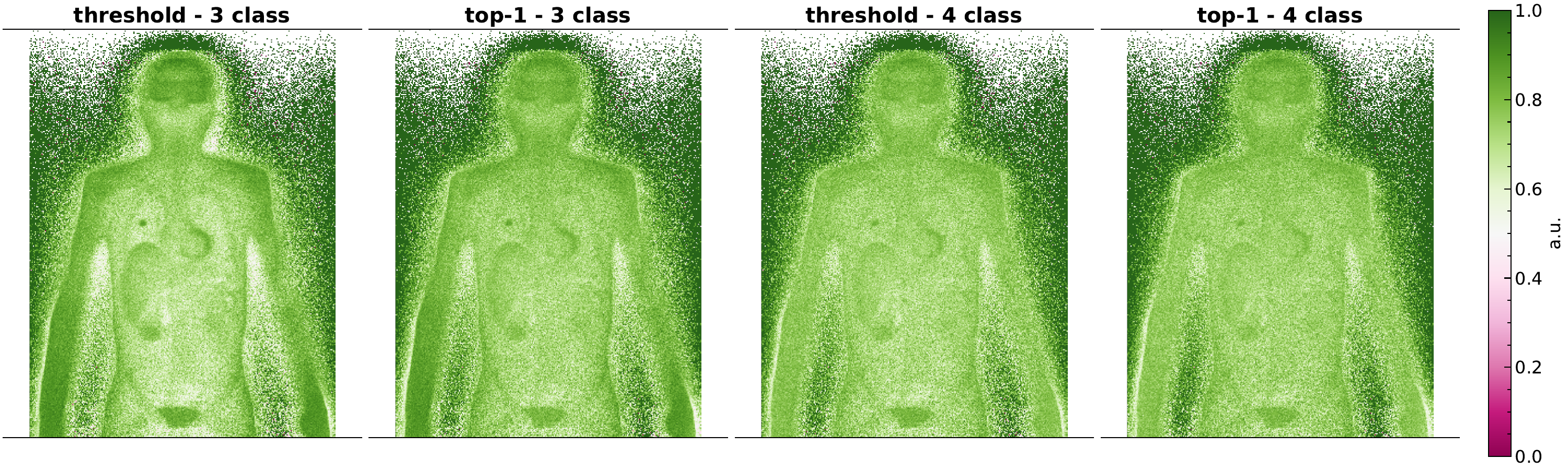}
  }\\
  \subfloat[]{
    \includegraphics[width=\linewidth]{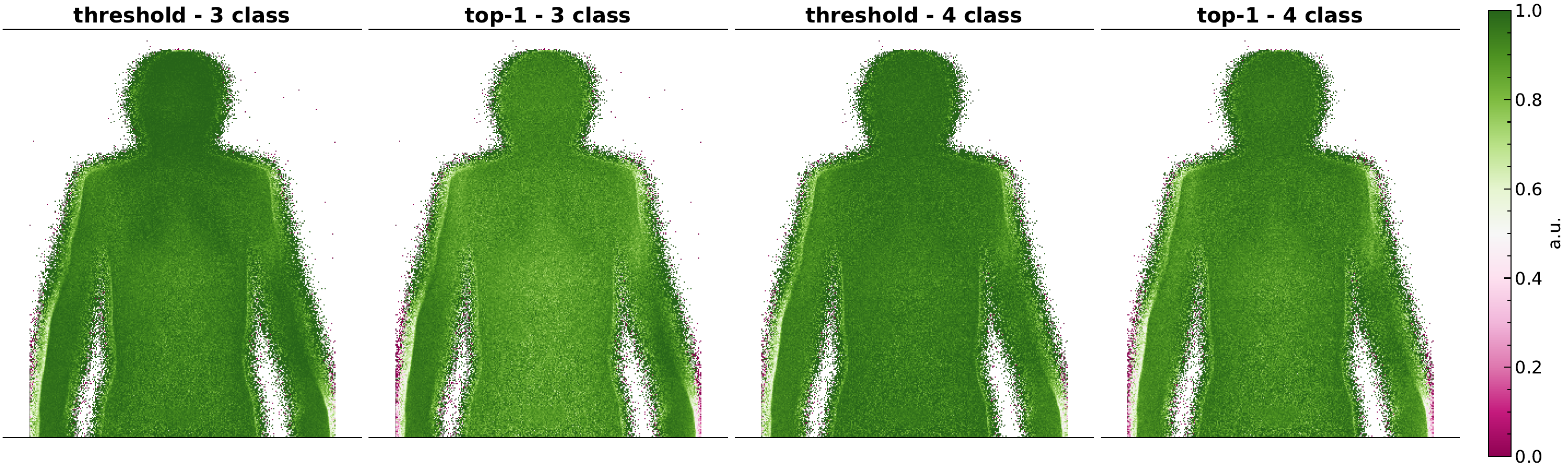}
  }
}
\caption{\label{Fig:metric_maps_top1_vs_threshold}Comparison of spatial distribution of the Accuracy (a) and TPR (b) metrics between {\it top-1} and {\it threshold}-based selection. Two XGB models are considered: XGB-3C-6F - an XGB model trained to classify data into 3 classes based on 6 features, and XGB-6F - an XGB model trained to classify data into 4 classes based on 6 features.}
\end{figure}

First, we evaluate the impact of the classification mode on the uniformity of the model's response. In Fig.~\ref{Fig:metric_maps_top1_vs_threshold} a comparison of accuracy and TPR maps is presented for the {\it threshold} and the {\it top-1} modes, for the \ac{XGBoost} model trained on the \ac{XCAT} sample. The {\it top-1} selection leads to more uniform model response, especially for the case where we restrict the classification problem to three classes by merging {\it DET} with {\it TRUE}, as the feature distributions of those classes largely overlap.

\begin{figure}[htbp]
\centering
  \subfloat[]{
    \includegraphics[width=\linewidth]{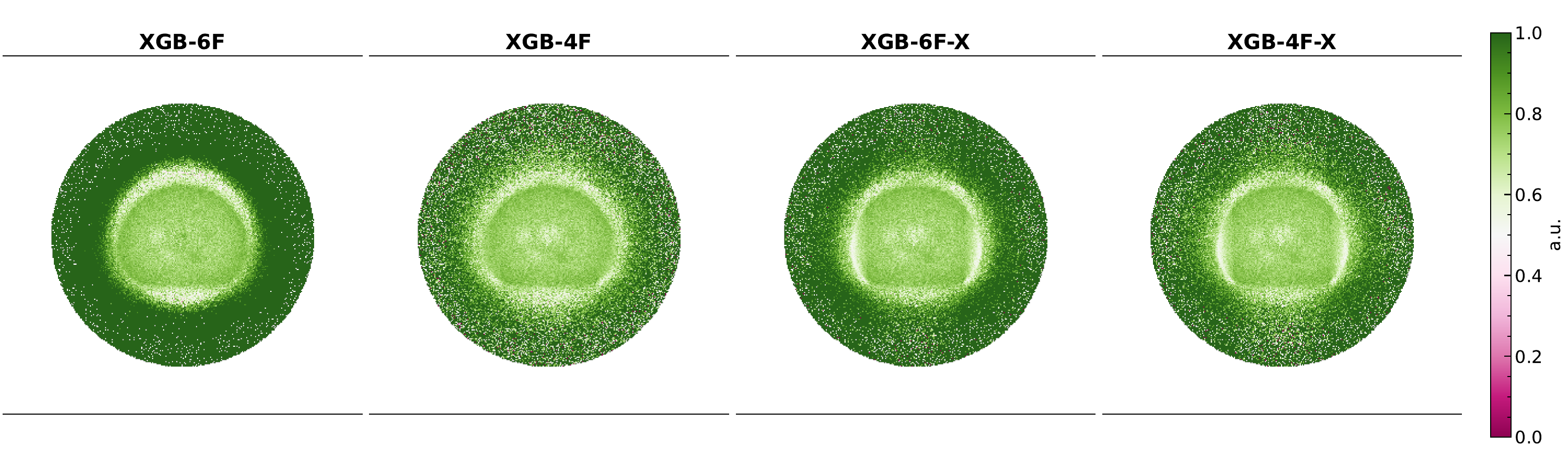}
  }\\
  \subfloat[]{
    \includegraphics[width=\linewidth]{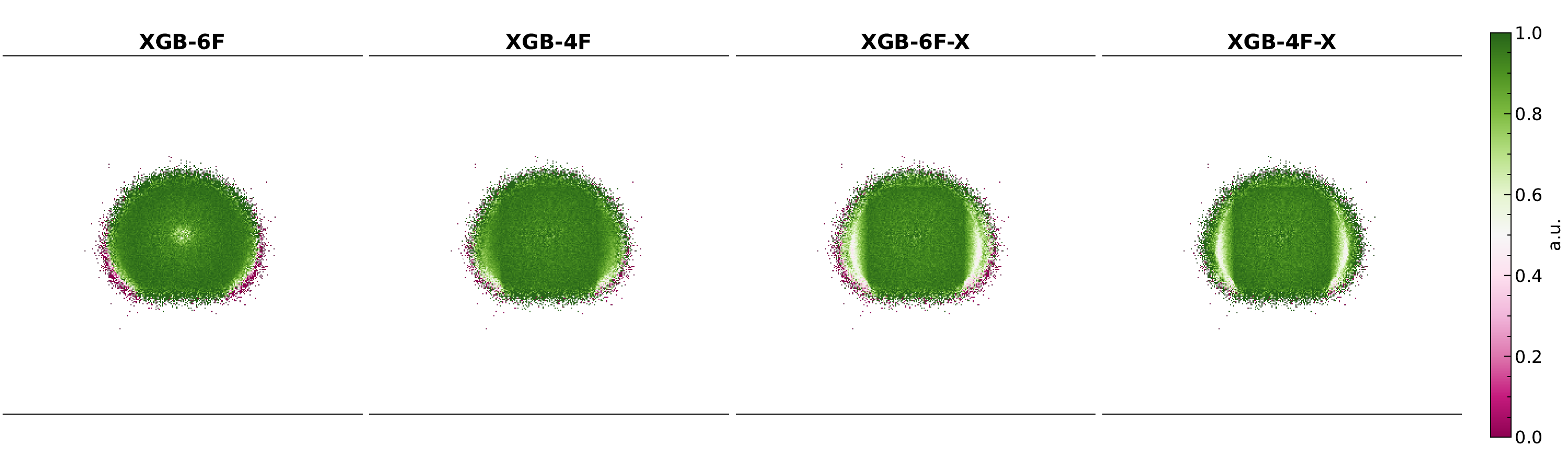}
  }\\
  \subfloat[]{
    \label{Fig:metric_maps_robustness_nema_PPV}
    \includegraphics[width=\linewidth]{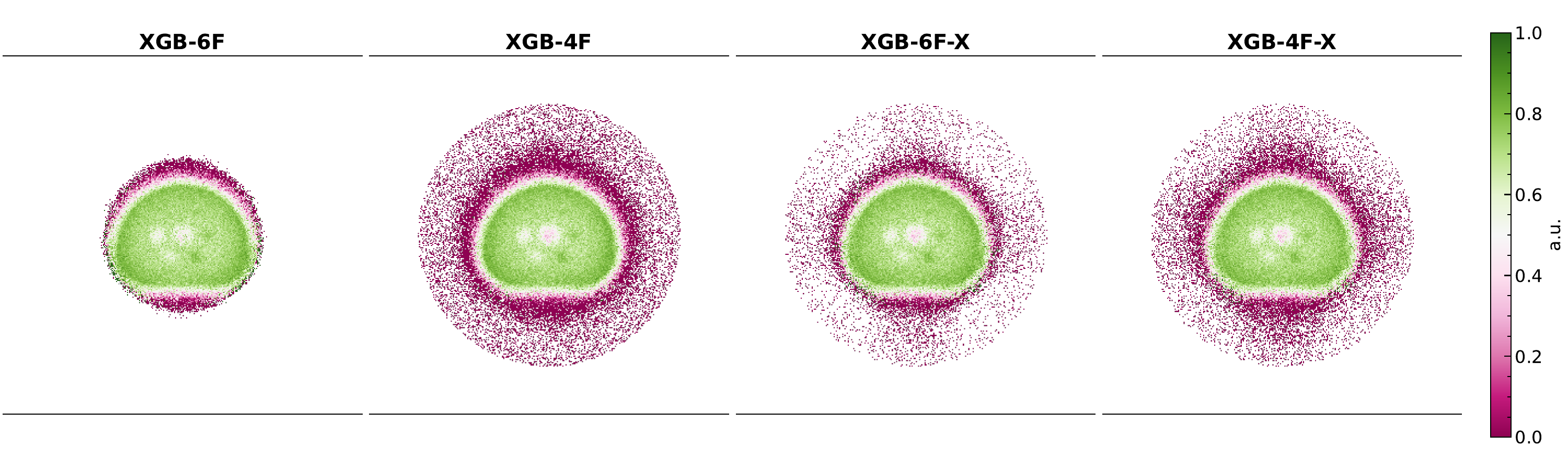}
}
\caption{\label{Fig:metric_maps_robustness_nema}Comparison of spatial distribution of the Accuracy (a), TPR (b) and PPV (c) metrics for the \ac{NEMA} phantom, between 4 XGB models: XGB-6F - a model trained on 6 features and \ac{NEMA} data; XGB-4F - a model trained on 4 features and \ac{NEMA} data; XGB-6F-X - a model trained on 6 features and \ac{XCAT} data; XGB-4F-X - a model trained on 4 features and \ac{XCAT} data.}
\end{figure}

We compare the impact of feature set selection and out-of-distribution evaluation on the \ac{XGBoost} model response. First, the performance on the \ac{NEMA} sample is assessed for four model variants (Fig.~\ref{Fig:metric_maps_robustness_nema}): two trained on the \ac{NEMA} sample with 6 features (XGB-6F) and four features (XGB-4F); two trained on the \ac{XCAT} sample also with 6 features (XGB-6F-X) and 4 features (XGB-4F-X). We observe that the model trained on the \ac{NEMA} with 6 features uses the $deg2D$ feature for a hard cylindrical rejection cut that tightly encompasses the phantom geometry. Another effect is visible around the lung insert: the model once again uses the $deg2D$ feature for positional classification of the coincidences in this region, while it improves the accuracy and PPV metrics, a clear degradation in TPR is observed. The variant trained on the \ac{XCAT} sample does not use such a hard threshold; nevertheless, the cylindrical shape with increased PPV metric is visible. Models trained on \ac{XCAT} data exhibit lower accuracy and TPR metrics, when evaluated on the \ac{NEMA} data, on the phantom extremes along the horizontal axis. We hypothesise that this is an effect of the \ac{XCAT} shape in which the only region that is close to the scanner main axis with unobstructed horizontal view is the phantom head. For the torso, the \ac{LOR}s originating from a similar region have an increased probability of crossing the phantom arms. This is not the case for the head; however, the horizontal extent of the \ac{XCAT} phantom in this region is smaller than for the \ac{NEMA} phantom and coincides with the region of high accuracy in the cross-inference spatial quality map.

\begin{figure}[htbp]
\centering {
  \subfloat[]{
    \includegraphics[width=\linewidth]{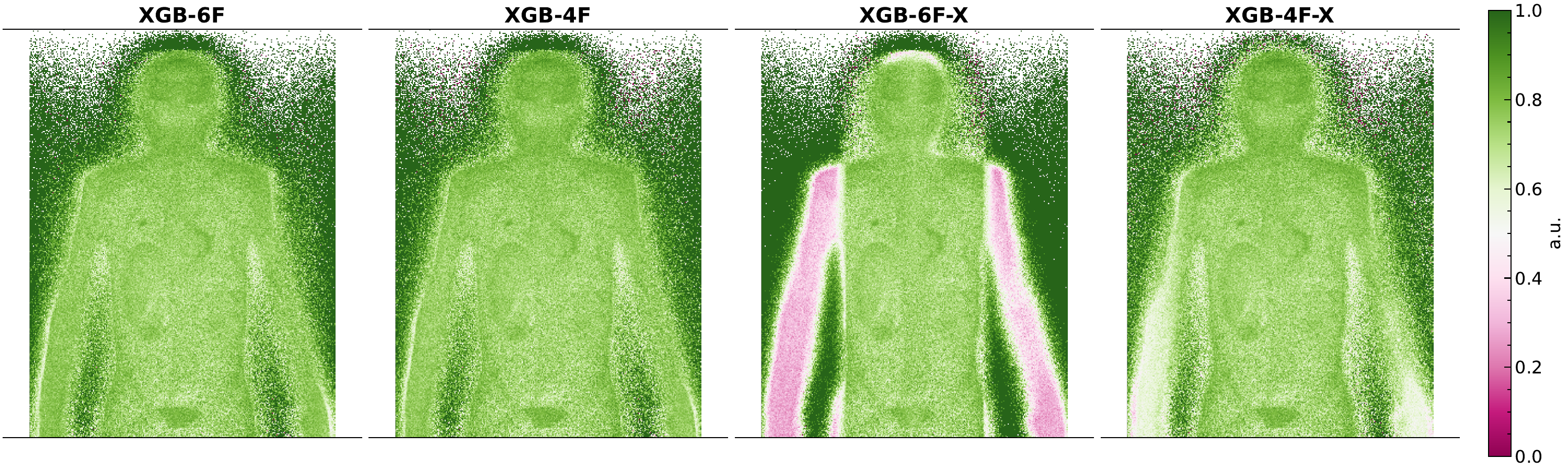}
  }\\
  \subfloat[]{
    \includegraphics[width=\linewidth]{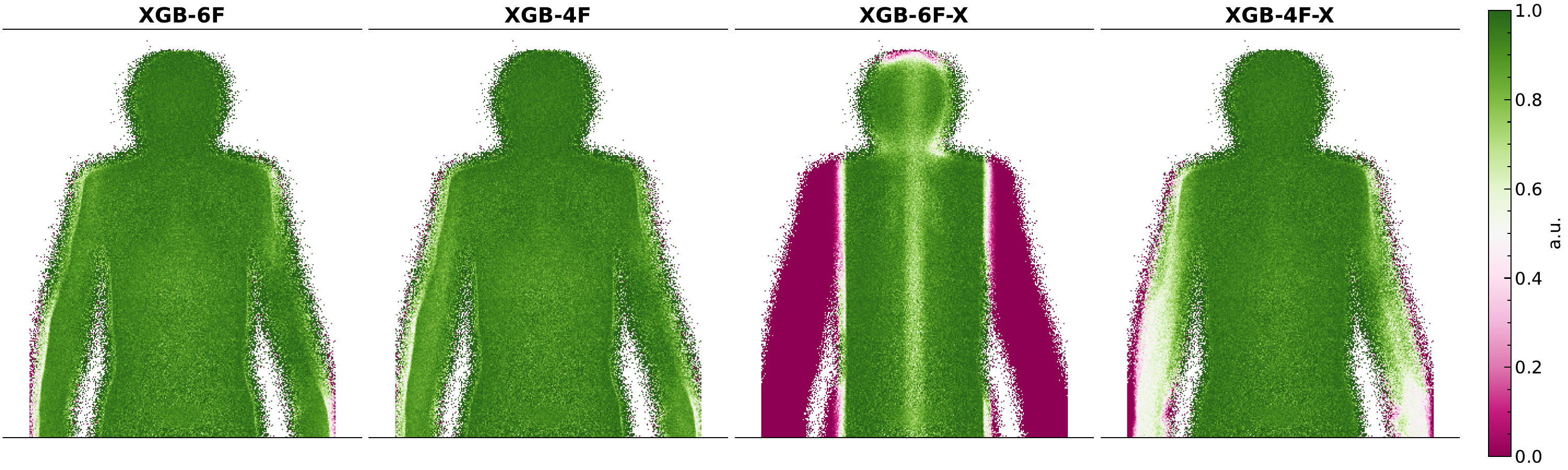}
  }\\
  \subfloat[]{
    \includegraphics[width=\linewidth]{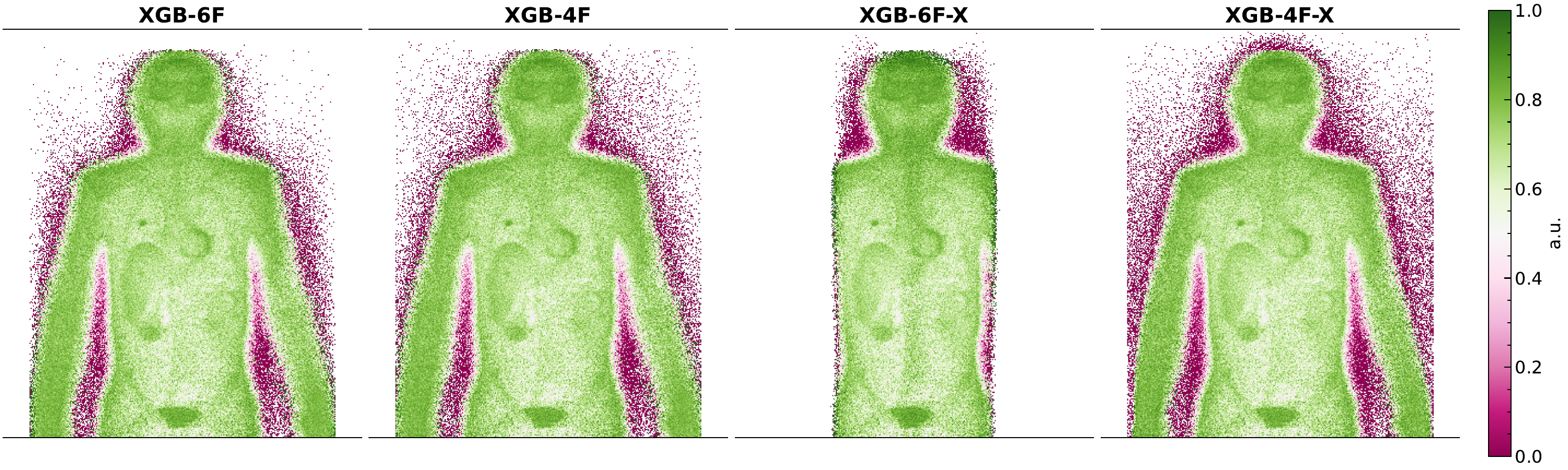}
  }
}
\caption{\label{Fig:metric_maps_robustness_xcat}Comparison of spatial distribution of the Accuracy (a), TPR (b) and PPV (c) metrics for the \ac{XCAT} phantom, between 4 XGB models: XGB-6F - a model trained on 6 features and \ac{XCAT} data; XGB-4F - a model trained on 4 features and \ac{XCAT} data; XGB-6F-X - a model trained on 6 features and \ac{NEMA} data; XGB-4F-X - a model trained on 4 features and \ac{NEMA} data.}
\end{figure}

Next, in the same way, we evaluate the \ac{XGBoost} model response uniformity on the \ac{XCAT} sample (Fig.~\ref{Fig:metric_maps_robustness_xcat}). First, we do not observe significant differences between the XGB-6F and the XGB-4F variants trained on the evaluated sample. The spatial diversity of the phantom prevents the model trained on a richer feature set from utilising simple geometrical shortcuts. On the other hand, the observed hard cuts for the 6-feature variant trained on the \ac{NEMA} sample (XGB-6F-X) are clearly visible. The effect is largely diminished in the four-feature variant (XGB-4F-X), with only the phantom's hands affected. This is most likely an effect of the $dt$ feature, which, while highly effective at rejecting the {\it ACC} coincidences, does encode the distance from the scanner main axis.

\section{Discussion}
\label{sec:Discussion}
\subsection{Feature importance interpretation} 
\label{sec:features}
The permutation importance analysis shows that the photon time difference dt and the attenuation factor AF are the two most discriminating features across all model architectures and simulated phantom setups.
This can be clearly interpreted in the context of the properties of background classes. Accidental coincidences are formed, by definition, from uncorrelated annihilation events, and the time difference between the two detected photons is therefore uncorrelated with the scanner geometry. This results in a broader $dt$ distribution for accidental pairs than for true coincidences, where the time difference is constrained by the finite size of the emission region and the scanner time resolution. Therefore, this feature provides direct, geometry-independent discrimination against the {\it ACC} class. The energy-based features carry limited discriminative power for the {\it ACC} class by construction; e.g. accidental pairs are not expected to differ systematically from the true coincidences in their deposited energy distributions, as both consist of unscattered 511~keV photons in the first approximation.
The attenuation factor encodes the total material traversed along the \ac{LOR}. For phantom-scattered photons, the \ac{LOR} is deflected; therefore, one can expect that the reconstructed line will, in general, traverse a different amount of attenuating material than a true coincidence originating from the same region. This makes the AF feature sensitive to the {\it PHT} class. Its advantage over the energy-based features $eSum$ and $eDiff$ suggests that spatial mislocation is a more reliable signature than energy deposition in the considered scenarios.

The relative importance of the secondary features, namely $eSum$ and $deg2D$, shows a dataset-dependent behaviour.
For the \ac{NEMA} IEC phantom, $deg2D$ gains importance relative to $AF$. We interpret it as the simplified, approximately cylindrical geometry of that phantom. The opening angle provides a way to encode the geometric cut separating in-phantom from out-of-phantom \ac{LOR}s.  This behaviour is consistent with the spatial quality maps discussed in Section~\ref{subsec:results_quality_maps}, where the 6-feature model trained on \ac{NEMA} IEC phantom applies a hard cylindrical selection driven by $deg2D$. Therefore, it can be interpreted as a kind of geometry memorisation that directly explains the degraded cross-phantom generalisation of the 6-feature model discussed in Section~\ref{subsec:results_robustness_test}.

\subsection{Robustness and generalization} 
\label{sec:robustness}

The cross-phantom results clearly show that model generalisation depends on the choice of feature set.
The 6-feature variant, which incorporates the topology-dependent variables $deg2D$ and $lorL$, suffers a substantially larger loss in accuracy when evaluated on out-of-distribution data than the 4-feature variant.
For the \ac{XGBoost} model trained on \ac{NEMA} and evaluated on the geometrically more complex \ac{XCAT} phantom, the accuracy loss amounts to $-0.130$ for the 6-feature variant, compared to $-0.044$ for the 4-feature one. 
This difference is physically expected: $deg2D$ and $lorL$ encode properties of the \ac{LOR} that are sensitive to the phantom shape and the position within the scanner \ac{FOV}. A model trained on a centrally positioned \ac{NEMA} phantom will learn decision boundaries in the topological feature space directly connected to the geometry. Therefore, their application to another anatomically complex phantom, such as \ac{XCAT}, becomes inappropriate. 
This observation is confirmed by the spatial quality maps obtained. The 6-feature model trained on the \ac{NEMA} dataset and evaluated on \ac{XCAT} introduces a hard cylindrical cut that removes a large portion of the \ac{XCAT} phantom, as discussed in section~\ref{subsec:results_quality_maps}. On the other hand, the 4-feature variant does not exhibit this sharp cut, and spatial degradation is greatly reduced. We attribute it to the removal of topology variables from the feature set; the remaining geometric sensitivity reflects the correlation of $dt$ with the distance from the scanner axis.

It is worth noting that the generalisation performance is asymmetric with respect to the direction of cross-inference. Models trained on the more complex \ac{XCAT} and evaluated on \ac{NEMA} suffer smaller accuracy losses than the reverse. In this case, a degradation in accuracy by $-0.010$ and $-0.018$ is observed for the 6-feature and 4-feature variants, respectively. However, we do observe two adverse effects: first, the 6-feature variant has a visible tendency to apply a circular cut in the transverse plane, visible in Fig.~\ref{Fig:metric_maps_robustness_nema_PPV} - a result of the inclusion of the $deg2D$ feature. In the evaluated setup, this does not lead to performance degradation as the geometry of the small phantom is largely cylindrical. Second, both variants result in degraded accuracy at the NEMA IEC phantom's extreme horizontal edges, indicating a mismatch between the clean geometry of the smaller phantom, which mimics the patient's torso, and the sophisticated phantom that includes the patient's arms. The results indicate that a careful selection of the training data phantom is crucial, with the best approach being a mixture of diverse phantoms. 

For the {\it top-1} selection mode, our best models achieve accuracies of up to $0.738$ and $0.691$ with an MCC metric of 0.60 and 0.51 for the NEMA IEC and \ac{XCAT} phantoms, respectively, outperforming the geometry-based cut baseline.
However, examination of the spatial quality maps and confusion matrices reveals an important limitation: the classifier's background rejection power is highest for the out-of-phantom coincidences and regions where \ac{LOR}s do not traverse the full phantom cross-section, resulting in a smaller scatter fraction. This is physically understood: phantom-scattered photons, especially at small angles, share kinematic properties similar to those of true coincidences, and the features available in this study provide limited information for distinguishing them from true coincidences.

We therefore conclude that event-by-event ML classification with a minimal low-level feature set represents a viable, geometry-agnostic alternative to traditional cut-based coincidence rejection, but does not on its own constitute a complete solution to the background problem in LAFOV PET. Meaningful further suppression of in-phantom scattered background will likely require incorporating additional discriminating information — such as Time-of-Flight consistency, higher-level topological features computed with respect to an estimated patient density map, or integration with iterative reconstruction — as well as evaluation on a broader range of clinical activity distributions. These directions define the natural scope of future work building on the foundation established here.

\section{Conclusions}
\label{sec:Conclusions}

We have presented a systematic study of event-by-event coincidence classification using machine learning as a pre-reconstruction background-rejection strategy for \ac{LAFOV} \ac{PET} scanners, evaluated via \ac{MC} simulations of the Siemens Biograph Vision Quadra with \ac{NEMA} and \ac{XCAT} phantoms. Three classifier architectures were compared:  \ac{XGBoost}, \ac{AdaBoost}, and \ac{MLP} across two feature sets and two phantom geometries, with generalisation assessed through cross-phantom inference. 

Our central finding is that neither using a compact 4-feature representation that excludes topological variables nor using a more complex phantom as training data guarantees high model performance on out-of-distribution data. A careful selection of the training features is critical as it can introduce non-trivial biases and provide an easy path for the model to shortcut a solution. It is common in the literature to find studies in which a model is trained on the largest available phantom that roughly covers the region of interest; for example, in~\cite{oliver_application_2013}, such an approach can lead to nontrivial model-deficiency regions. A recommended approach is to use a diverse set of phantoms during training to represent expected variations in the target data.

The spatial quality map analysis, which exploits the one-to-one correspondence between coincidence events and reconstructed voxels provided by direct \ac{TOF} reconstruction, reveals localised performance patterns that global metrics alone would not detect. This methodology is proposed as a standard complement to aggregate performance evaluation in future studies of event-by-event ML filtering. In particular, they offer an important insight for the crucial task of feature and training data selection.

The primary performance boundary of the current approach is the phantom-scattered coincidence class, which constitutes approximately 25\% of all events and remains difficult to reject with the considered feature set due to the overlap in observable properties with true coincidences. In this work, we establish a baseline and methodology for future work on extended event-by-event methods that will incorporate additional discriminating information.

%
% Each of the commands below will create an unnumbered section with the appropriate heading.
% Remove any sections that are not relevant for your article.
% All sections except suppdata will be removed if the [anonymous] option is used.
% See iopjournal-guidelines.pdf for more information.
%

\ack{
The authors acknowledge the technical support of A. Spirzewska.
This work was completed with resources provided by the Świerk Computing Centre at the National Centre for Nuclear Research.
We gratefully acknowledge Polish high-performance computing infrastructure PLGrid (HPC Center: ACK Cyfronet AGH) for providing computer facilities and support within computational grant no. PLG/2024/017403.
\ifarxiv
The „IMPET – Industrial Multiphoton PET Tomography FENG.02.02-IP.05-0152/23” project is carried out within the FIRST TEAM FENG programme of the Foundation for Polish Science co-financed by the European Union under the European Funds for Smart Economy 2021-2027 (FENG).
\fi
}

\ifarxiv
% skip
\else
\funding{
The „IMPET – Industrial Multiphoton PET Tomography FENG.02.02-IP.05-0152/23” project is carried out within the FIRST TEAM FENG programme of the Foundation for Polish Science co-financed by the European Union under the European Funds for Smart Economy 2021-2027 (FENG).
}
% This section is a list of funder names and grant numbers

%\roles{Sample text inserted for demonstration.}
% List author names and the contributions made to the article, using terms from the NISO Contributor Roles Taxonomy (CRediT) https://credit.niso.org

%\data{Sample text inserted for demonstration.}
% For more information on IOP Publishing's research data policy see: https://publishingsupport.iopscience.iop.org/questions/research-data/

%\suppdata{Sample text inserted for demonstration.}
\fi

%\listoftodos

\section*{References}

\clearpage
\bibliography{ML_article}

%%Appendices
\appendix

%%Appendix: Hyperparams
\section{Tuning of model hyperparameters}
\label{appendix:hyperparamaters}

\begin{table*}[!t]
\centering
\begin{tabular}{| c | c c c | c c |} 
 \hline
 model & hyperparameter & range & prior & best NEMA & best \ac{XCAT} \\ 
 \hline\hline
 \ac{AdaBoost} & learning rate & $(1.0\times10^{-2}, 1.0)$ & log-uniform & 0.35 & 0.05 \\ 
          & max depth     & $(1, 30)$                 & uniform     & 3    & 7   \\
          & estimators    & $(2, 200)$                & log-uniform & 200  & 200 \\
 \hline
 \ac{MLP} & learning rate & $(1.0\times10^{-2}, 1.0)$ & log-uniform & 0.2 & 0.2 \\
                & layer size    & $(2, 300)$                & log-uniform & 205 & 300 \\
                & layer count   & $(2, 10)$                 & uniform     & 2   & 2   \\
                & dropout       & $(0.0, 0.9)$              & uniform     & 0.0 & 0.0 \\
                & batch size    & $(32, 4096)$              & log-uniform & 223 & 555 \\
 \hline
 \ac{XGBoost} & learning rate    & $(1.0\times10^{-3}, 1.0\times10^{-1})$ & log-uniform & 0.1 & 0.08 \\
         & max depth        & $(1, 10)$                              & uniform     & 8   & 9 \\
         & gamma            & $(1.0\times10^{-6}, 1.0\times10^{4})$  & log-uniform & $1.0\times10^{-6}$ & 0.05 \\
         & subsample        & $(0.5, 1.0)$                           & uniform     & 0.79 & 0.58 \\
         & colsample bytree & $(0.5, 1.0)$                           & uniform     & 1.0 & 0.98 \\
         & min child weight & $(1, 5)$                               & uniform     & 5   & 2 \\
         & estimators       & $(2, 200)$                             & log-uniform & 200 & 175 \\
         & alpha            & $(1.0\times10^{-6}, 1.0\times10^{3})$  & log-uniform & $1.0\times10^{-6}$ & $5.0\times10^{-6}$ \\
         & lambda           & $(1.0\times10^{-6}, 1.0\times10^{3})$  & log-uniform & $1.0\times10^{-6}$ & $2.0\times10^{-6}$ \\
 \hline
\end{tabular}
\caption{Model hyperparameters evaluated using Bayesian Optimisation, together with the best values obtained for each dataset.}
\label{table:hyperparameters_ranges}
\end{table*}

\begin{figure}[!t]
\centering
  \subfloat[]{
    \includegraphics[width=0.33\linewidth]{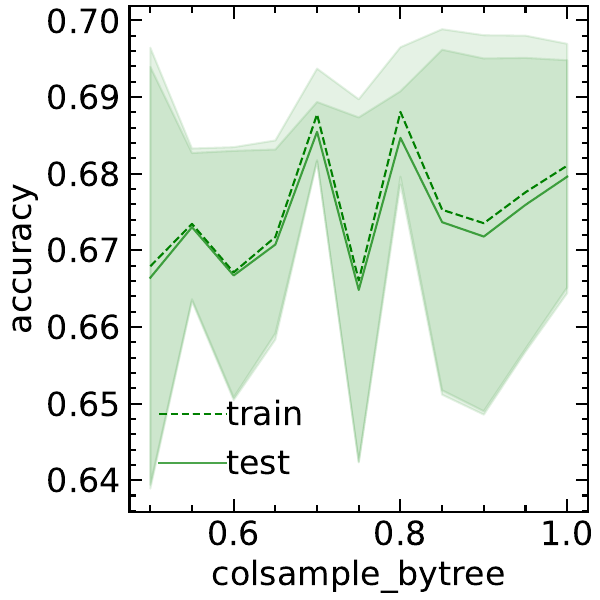}
  }
  \subfloat[]{
    \includegraphics[width=0.33\linewidth]{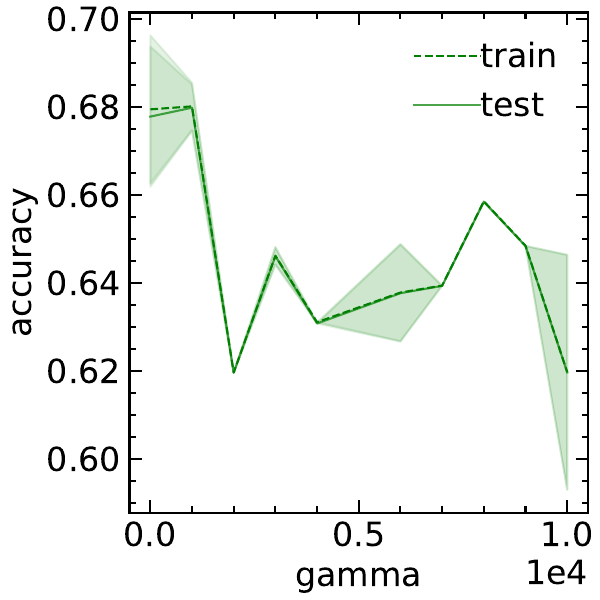}
  }
  \subfloat[]{
    \includegraphics[width=0.33\linewidth]{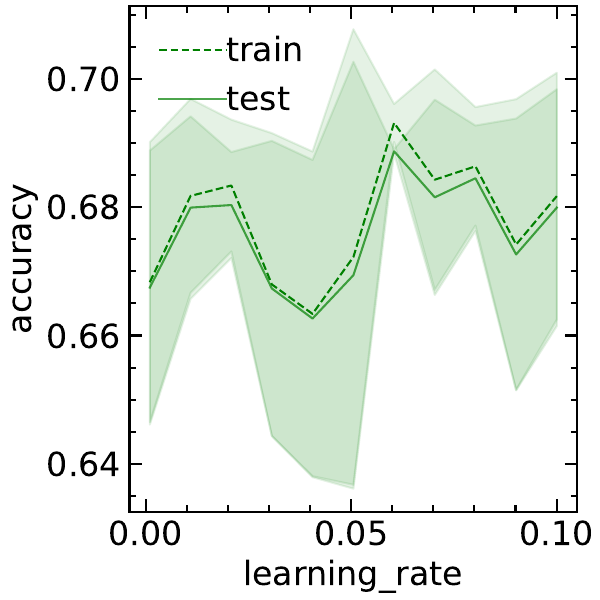}
  }\\
  \subfloat[]{
    \includegraphics[width=0.33\linewidth]{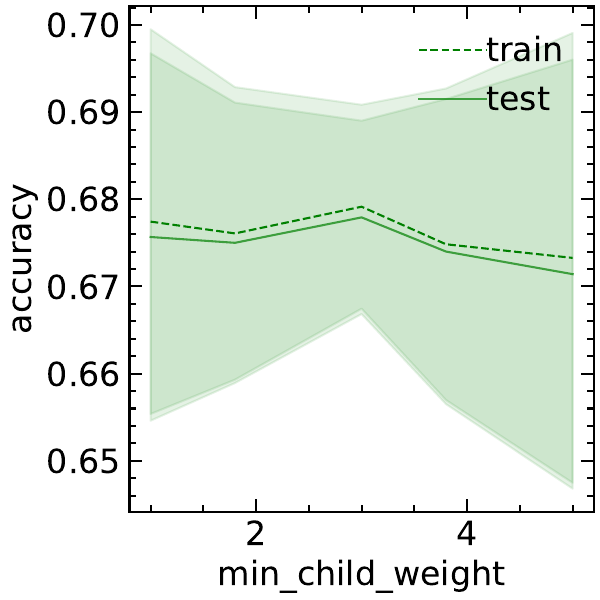}
  }
  \subfloat[]{
    \includegraphics[width=0.33\linewidth]{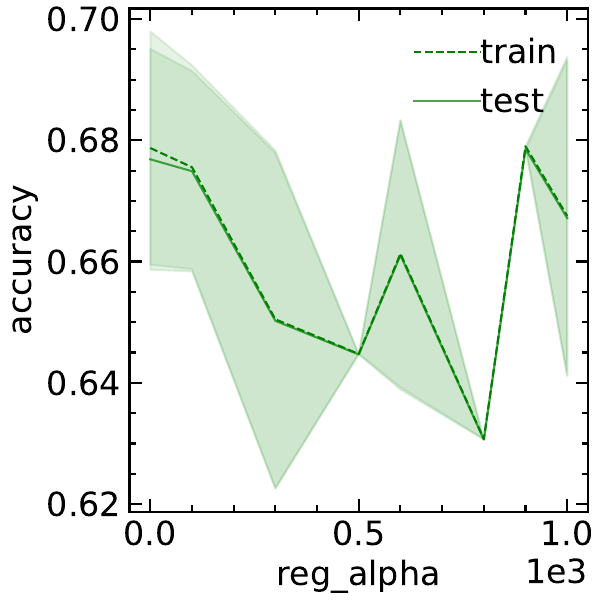}
  }
  \subfloat[]{
    \includegraphics[width=0.33\linewidth]{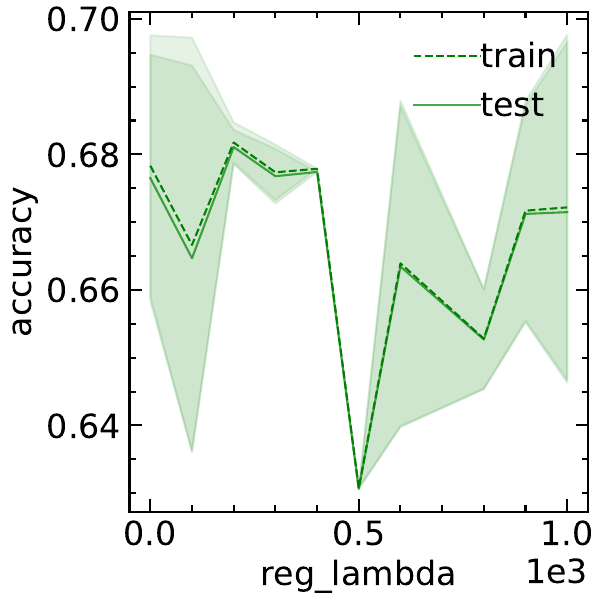}
  } \\
  \subfloat[]{
    \includegraphics[width=0.33\linewidth]{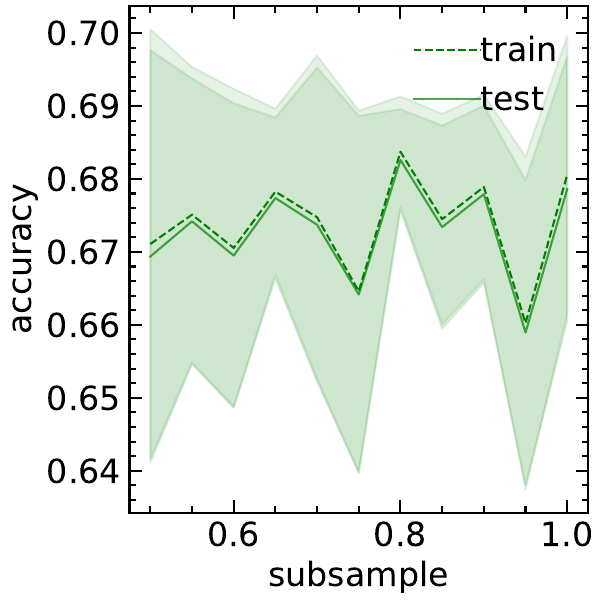}
  }
\caption{Distribution of accuracy as a function of regularisation hyperparameters values for the 6-feature \ac{XGBoost} model trained on the \ac{XCAT} sample.}
\label{Fig:hyperparam_xgb_6f_all}
\end{figure}

\begin{figure}[!t]
\centering
  \subfloat[]{
    \includegraphics[width=0.33\linewidth]{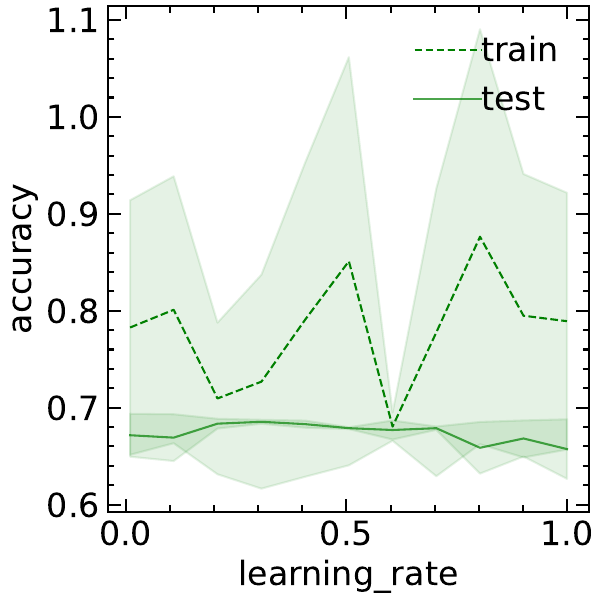}
  }
  \subfloat[]{
    \includegraphics[width=0.33\linewidth]{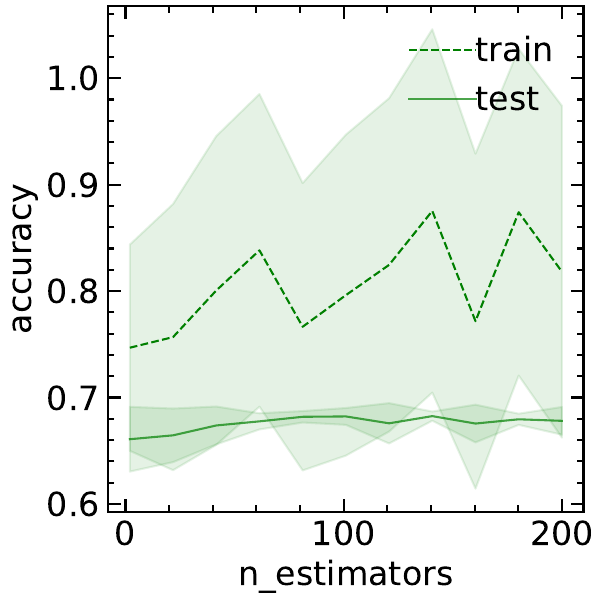}
  }
  \subfloat[]{
    \includegraphics[width=0.33\linewidth]{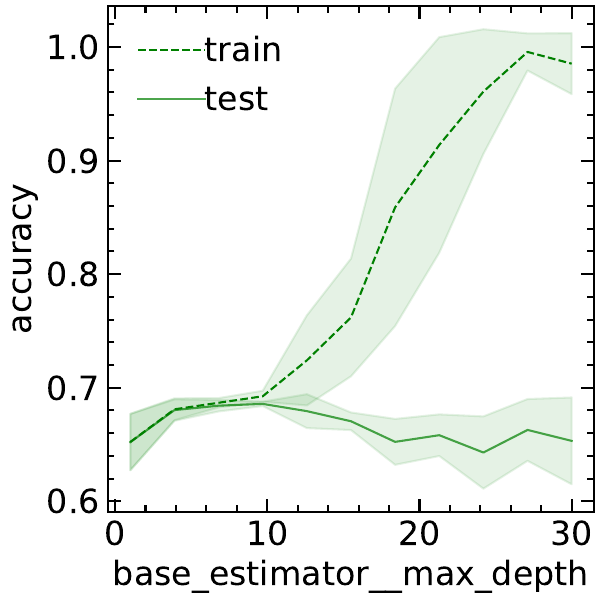}
  }
\caption{Distribution of accuracy as a function of hyperparameter values for the 6-feature \ac{AdaBoost} model trained on the \ac{XCAT} sample.}
\label{Fig:hyperparam_ada_6f_all}
\end{figure}

\begin{figure}[!t]
\centering
  \subfloat[]{
    \includegraphics[width=0.33\linewidth]{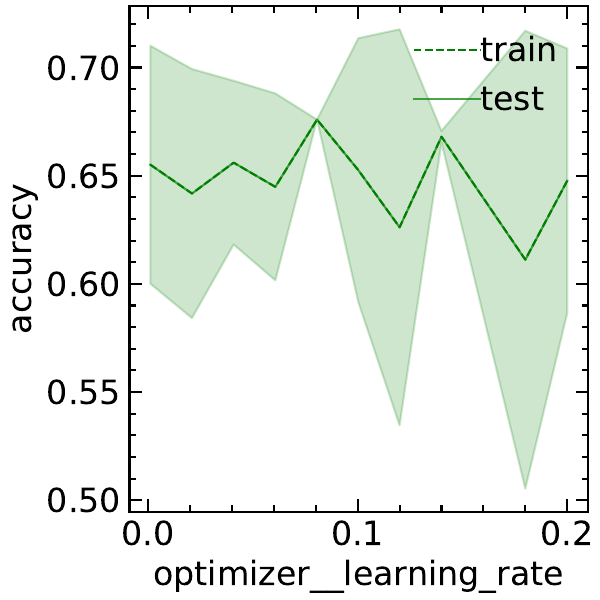}
  }
  \subfloat[]{
    \includegraphics[width=0.33\linewidth]{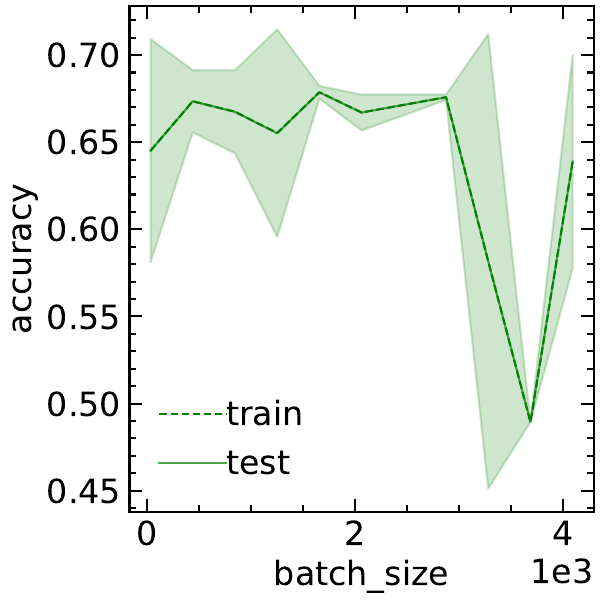}
  }
  \subfloat[]{
    \includegraphics[width=0.33\linewidth]{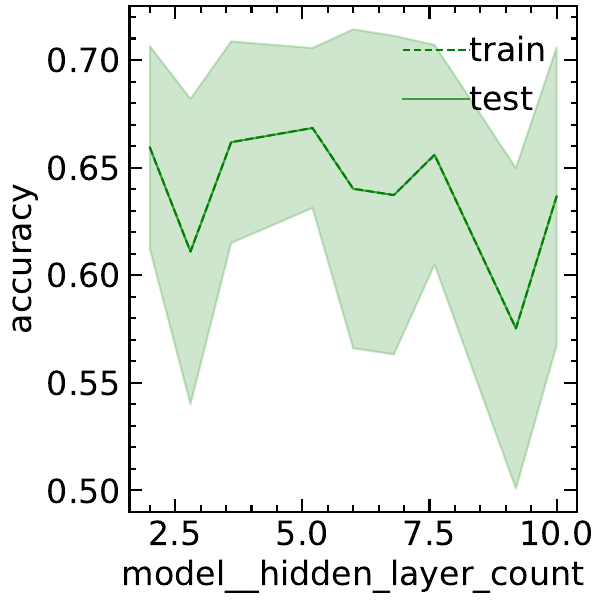}
  }\\
  \subfloat[]{
    \includegraphics[width=0.33\linewidth]{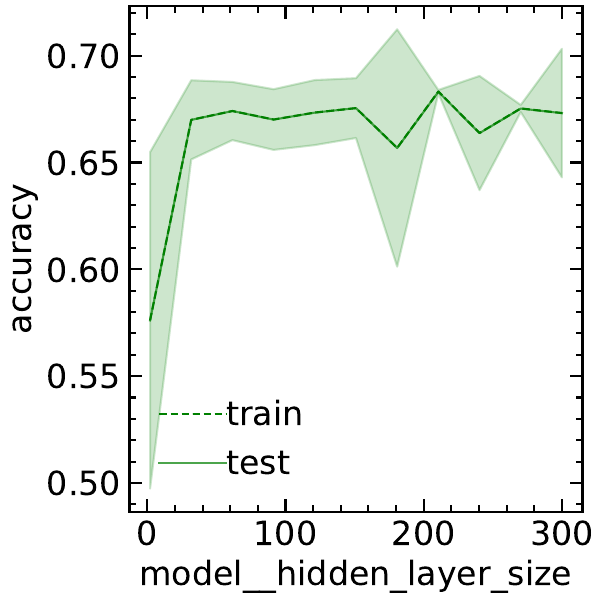}
  }
  \subfloat[]{
    \includegraphics[width=0.33\linewidth]{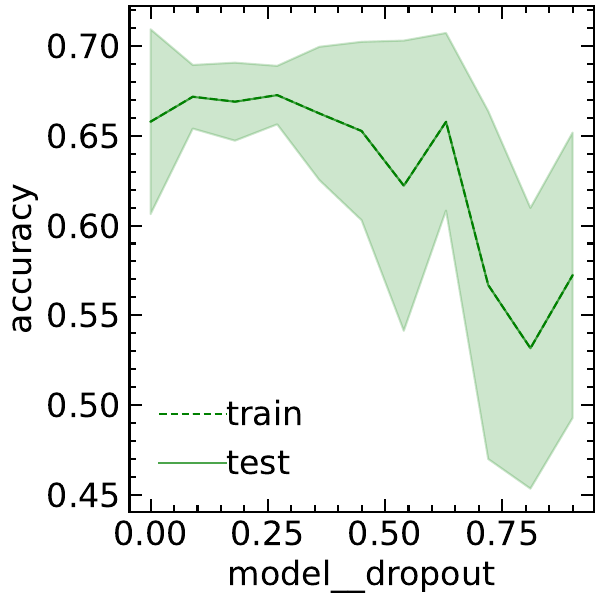}
  }
\caption{Distribution of accuracy as a function of hyperparameter values for the 6-feature NN model trained on the \ac{XCAT} sample.}
\label{Fig:hyperparam_nn_6f_all}
\end{figure}

As described in section~\ref{subsec:methods_hyperparameters} the model's hyperparameters are tuned using Bayesian Optimisation. Both the search ranges and the best values for each hyperparameter are presented in the Table~\ref{table:hyperparameters_ranges}. In the table the best values obtained for the \ac{NEMA} and \ac{XCAT} samples are compared. It can be observed that most of them have comparable values for the two samples. The accuracy dependence on the hyperparameter values is presented in the Figures~\ref{Fig:hyperparam_xgb_6f_example}, \ref{Fig:hyperparam_xgb_6f_all}, \ref{Fig:hyperparam_ada_6f_all}, and \ref{Fig:hyperparam_nn_6f_all}, for the \ac{XGBoost}, \ac{AdaBoost}, and \ac{NN} models, respectively. For brevity it is only presented for the \ac{XCAT} sample.

%%Appendix: Feature importance
\section{Feature importance}
\label{appendix:feature_importance}

\begin{figure}[htb]
\centering {
  \subfloat[]{
    \includegraphics[width=0.49\linewidth]{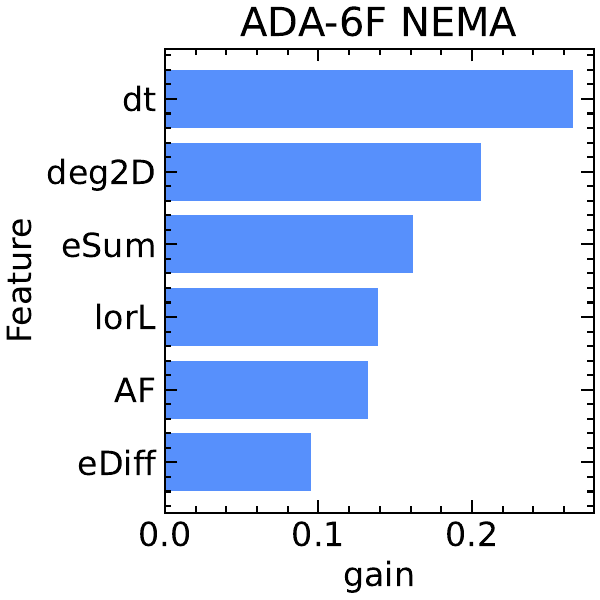}
  }
  \subfloat[]{
    \includegraphics[width=0.49\linewidth]{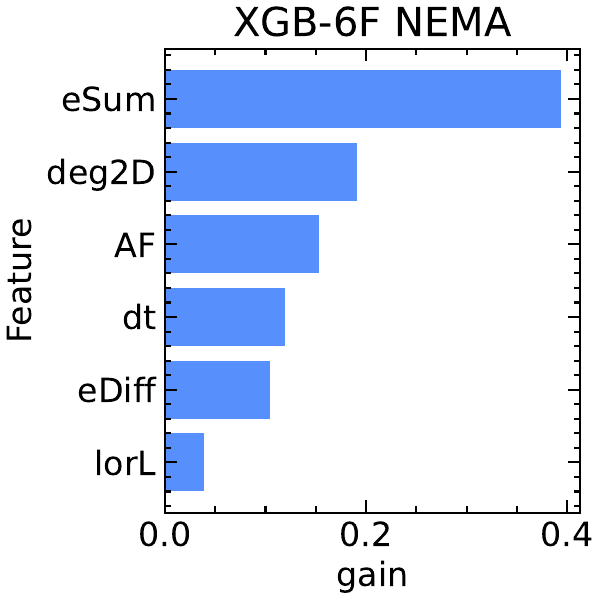}
  }\\
  \subfloat[]{
    \includegraphics[width=0.49\linewidth]{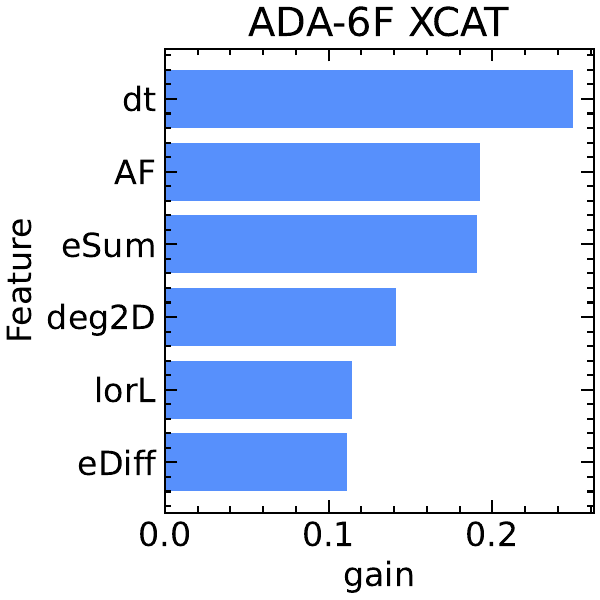}
  }
  \subfloat[]{
    \includegraphics[width=0.49\linewidth]{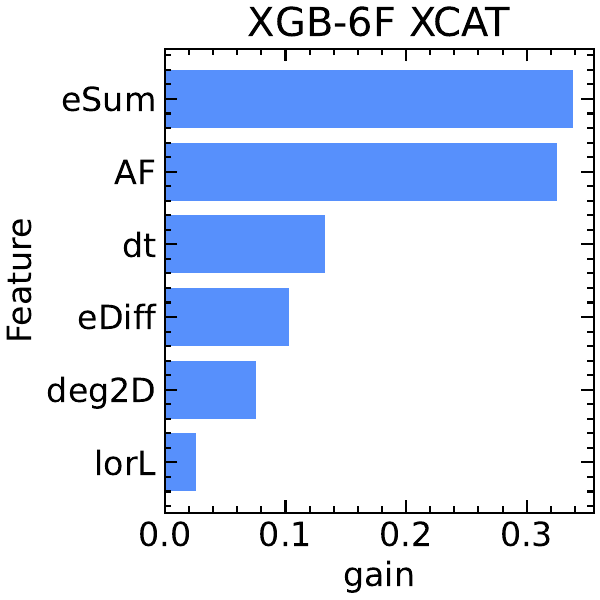}
  }
}
\caption{\label{fig:feature_importance}Feature importance for the \ac{AdaBoost} (left column) and \ac{XGBoost} (right column) models trained on the \ac{NEMA} (top row) and \ac{XCAT} (bottom row) samples.}
\end{figure}

\begin{figure}[htb]
\centering {
  \subfloat[]{
    \includegraphics[width=0.33\linewidth]{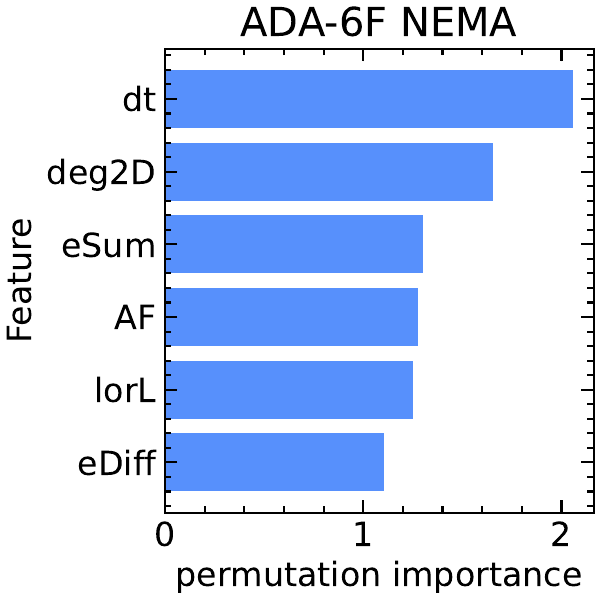}
  }
  \subfloat[]{
    \includegraphics[width=0.33\linewidth]{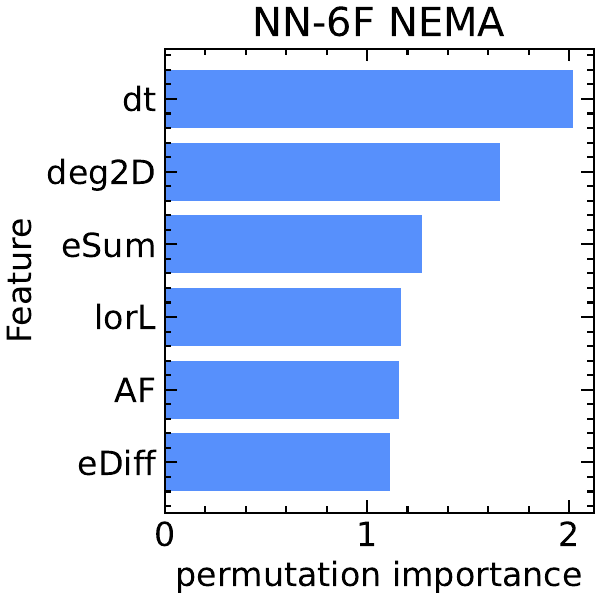}
  }
  \subfloat[]{
    \includegraphics[width=0.33\linewidth]{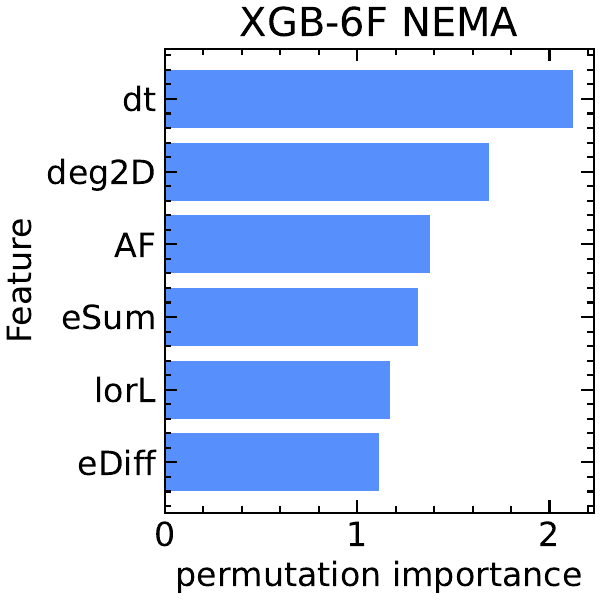}
  }\\
  \subfloat[]{
    \includegraphics[width=0.33\linewidth]{figures/ml/XCAT/ADA/PermutationFeaturesImportance.pdf}
  }
  \subfloat[]{
    \includegraphics[width=0.33\linewidth]{figures/ml/XCAT/NN/PermutationFeaturesImportance.pdf}
  }
  \subfloat[]{
    \includegraphics[width=0.33\linewidth]{figures/ml/XCAT/XGB/PermutationFeaturesImportance.pdf}
  }
}
\caption{\label{fig:permutation_feature_importance}Permutation feature importance for the \ac{AdaBoost} (left column), \ac{NN} (central column), and \ac{XGBoost} (right column) models trained on the \ac{NEMA} (top row) and \ac{XCAT} (bottom row) samples.}
\end{figure}

The feature importance, decribed in section~\ref{sec:features}, for the 6-feature \ac{XGBoost} and \ac{AdaBoost} models trained on both the \ac{NEMA} and \ac{XCAT} samples is presented in Fig.~\ref{fig:feature_importance}. For all models the permutation importance is presented in Fig.~\ref{fig:permutation_feature_importance}, also for both data samples.

\end{document}